\documentclass[10pt,twocolumn,letterpaper]{article}

\usepackage[pagebackref=true,breaklinks=true,colorlinks,bookmarks=false]{hyperref}

\usepackage{titlesec}
\usepackage{iccv}
\usepackage{epsfig}
\usepackage{graphicx}
\usepackage{amsmath}
\usepackage{amssymb}
\usepackage{amsthm}
\usepackage{mathtools}

\usepackage{algorithm}
\usepackage{algpseudocode}
\usepackage{cleveref}
\usepackage{xcolor}
\usepackage{enumitem}

\usepackage{newtxtext}
\usepackage{newtxmath}
\usepackage{adjustbox}
\usepackage{subcaption}

\newcommand{\wireframe}{\mathsf{W}}

\newcommand{\junction}{\mathbf{p}}

\newcommand{\vertexSet}{{\mathsf{V}}}
\newcommand{\edgeSet}{\mathsf{E}}

\newcommand{\bin}{b}
\newcommand{\location}{\mathbf{b}}
\newcommand{\locationA}{\mathbf{q}}

\renewcommand{\bin}{b}
\newcommand{\staticSet}{\mathbb{S}}
\newcommand{\dynamicSet}{\mathbb{D}}

\newcommand{\junctionMap}{J}
\newcommand{\offsetMap}{\mathbf{O}}

\newcommand{\junctionType}{t}

\newcommand{\threshold}{\vartheta}

\newcommand{\N}[1]{{\color{red}{#1}}}
\renewcommand{\N}[1]{{{#1}}}

\newcommand{\requireJunctype}[1]{}
\newcommand{\changeurlcolor}[1]{\hypersetup{urlcolor=#1}}

\titlespacing*{\paragraph}{0pt}{0.6ex}{1em}
\setitemize{itemsep=2pt,topsep=0pt,parsep=0pt,partopsep=0pt}

\DeclareMathOperator*{\argmin}{arg\,min}

\iccvfinalcopy %
\def\iccvPaperID{1586} %

\ificcvfinal\fi

\begin{document}

\title{End-to-End Wireframe Parsing}

\author{Yichao Zhou \\
UC Berkeley \\
\tt \changeurlcolor{black}\href{mailto:zyc@berkeley.edu}{zyc@berkeley.edu}
\and
Haozhi Qi\\
UC Berkeley \\
\tt \changeurlcolor{black}\href{mailto:hqi@berkeley.edu}{hqi@berkeley.edu}
\and
Yi Ma\\
UC Berkeley \\
\tt \changeurlcolor{black}\href{mailto:yima@berkeley.edu}{yima@berkeley.edu}
}

\maketitle

\setlength{\abovedisplayskip}{5pt}
\setlength{\belowdisplayskip}{5pt}
\addtolength{\textfloatsep}{-8pt}
\setlength{\belowcaptionskip}{-2pt}

\begin{abstract}

We present a conceptually simple yet effective algorithm to detect wireframes \cite{Huang:2018:LPW} in a given image. Compared to the previous methods \cite{Huang:2018:LPW,xue2018learning} which first predict an intermediate heat map and then extract straight lines with heuristic algorithms, our method is end-to-end trainable and can directly output a vectorized wireframe that contains semantically meaningful and geometrically salient junctions and lines. To better understand the quality of the outputs, we propose a new metric for wireframe evaluation that penalizes overlapped line segments and incorrect line connectivities. We conduct extensive experiments and show that our method significantly outperforms the previous state-of-the-art wireframe and line extraction algorithms \cite{Huang:2018:LPW,xue2018learning,von2010lsd}. We hope our simple approach can be served as a baseline for future wireframe parsing studies. Code has been made publicly available at \url{https://github.com/zhou13/lcnn}.

\end{abstract}

\section{Introduction} \label{sec:introduction}
Recent progress in object recognition \cite{krizhevsky2012imagenet, szegedy2015going, simonyan2014very, he2016deep} and large-scale datasets \cite{ILSVRC15, dai2017scannet, chang2015shapenet, armeni2017joint} has made it possible to recognize, extract, and utilize high-level geometric features or global structures of a scene for image-based 3D reconstruction. Unlike local features (SIFT \cite{lowe1999object}, ORB \cite{rublee2011orb}, etc.) used in conventional 3D reconstruction systems such as structure from motion (SfM) and visual SLAM, high-level geometric features provide more salient and robust information about the global geometry of the scene. This line of research has drawn interests on the exploration of extracting structures such as lines and junctions (wireframes) \cite{Huang:2018:LPW}, planes \cite{yang2018recovering,liu2018planenet}, surfaces \cite{groueix2018papier}, and room layouts \cite{zou2018layoutnet}.  %

\begin{figure}[t]
    \centering
    \begin{subfigure}[b]{0.42\linewidth}
    \frame{\includegraphics[width=0.99\linewidth]{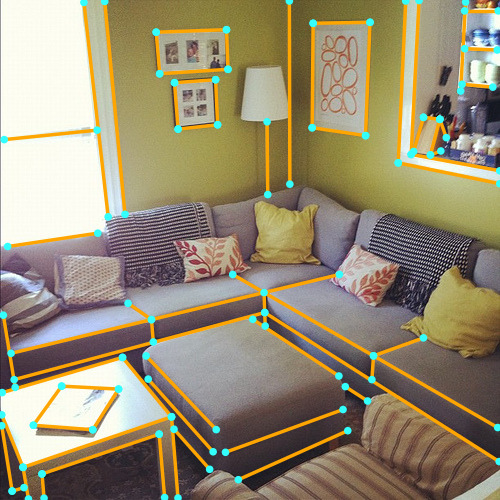}}
    \caption{Ground truth labels}
    \end{subfigure}
    \hspace{8pt}
    \begin{subfigure}[b]{0.42\linewidth}
    \frame{\includegraphics[width=0.99\linewidth]{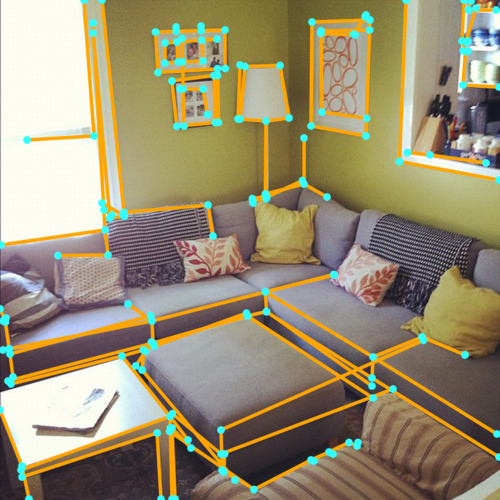}}
    \caption{Wireframe \cite{Huang:2018:LPW}}
    \end{subfigure}
    
    \vspace{5pt}

    \begin{subfigure}[b]{0.42\linewidth}
    \frame{\includegraphics[width=0.99\linewidth]{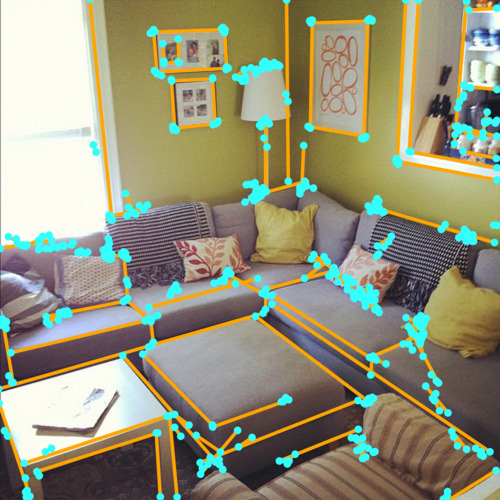}}
    \caption{AFM \cite{xue2018learning}}
    \end{subfigure}
    \hspace{8pt}
    \begin{subfigure}[b]{0.42\linewidth}
    \frame{\includegraphics[width=0.99\linewidth]{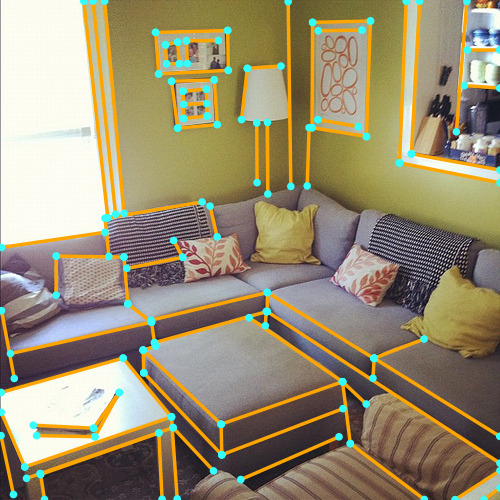}}
    \caption{Our proposed L-CNN}
    \end{subfigure}
    \caption{Demonstration of the wireframe representation of a scene and the results produced by Wireframe \cite{Huang:2018:LPW}, AFM \cite{xue2018learning}, and our proposed L-CNN.}
    \label{fig:teaser}
\end{figure}
Among all the high-level geometric features, straight lines and their junctions (together called a wireframe \cite{Huang:2018:LPW}) are probably the most fundamental elements that can be used to assemble the 3D structures of a scene. Recently, works such as \cite{Huang:2018:LPW} encourages the research of wireframe parsing by providing a well-annotated dataset, a learning-based framework, as well as a set of evaluation metrics. Nevertheless, existing wireframe parsing systems are intricate and still inadequate for complex scenes with complicated line connectivity. The goal of this paper is to explore a clean and effective solution to this challenging problem. 

Existing researches \cite{Huang:2018:LPW, xue2018learning} address the wireframe parsing problem with two stages. First, an input image is passed through a deep convolutional neural network to generate pixel-wise junction and line heat maps (or their variants \cite{xue2018learning}). After that, a heuristic algorithm is used to search through the generated heat map to find junction positions, vectorized line segments, and their connectivity. While these methods are intuitive and widely used in the current literature, their vectorization algorithms are often complex and rely on a set of heuristics, and thus sometimes lead to inferior solutions. Inspired by \cite{dai2016instance, he2017mask, girshick2015fast} in which the end-to-end pipelines outperform their stage-wise counterparts, we hypothesis that making wireframe parsing systems end-to-end trainable could also push the state-of-the-arts. Therefore, in this paper we address the following problem:
\begin{quote}
    {\em How to learn a vectorized representation of wireframes in an end-to-end trainable fashion?}
\end{quote}

To this end, we propose a new network called {\em L-CNN}, an algorithm that performs end-to-end wireframe parsing using a single and unified neural network. Our network can be split into four parts: a feature extraction backbone, a junction proposal module, and a line verification module bridged by a line sampling module. Taken an RGB image as the input, the neural network directly generates a vectorized representation without using heuristics. Our system is fully differentiable and can be trained end-to-end through back-propagation, enabling us to fully exploit the power of the state-of-the-art neural network architectures to parse the scenes.

Besides, current wireframe evaluation metrics treat a line as a collection of independent pixels, so it cannot take the correctness of line connectivity into consideration, as discussed in \Cref{sec:metric}. To evaluate such structural correctness of a wireframe, we introduce a new evaluation metric. Our new proposed metric uses line matching to calculate the precision and recall curves on vectorized wireframes.  We perform extensive experiments on wireframe datasets \cite{Huang:2018:LPW} and carefully do the ablation study on the effects of different system design choices.

\section{Related work} \label{sec:related}

\paragraph{Line Detection:} Line detection is a widely studied problem in computer vision. It aims to produce vectorized line representation from images. Traditional methods such as \cite{stephens1991probabilistic,von2010lsd} detect lines based on local edge features. Recently, \cite{xue2018learning} combines the deep learning-based features with the line vectorization algorithm from \cite{von2010lsd}. Unlike the wireframe representation, traditional line detection algorithms do not provide the information about junctions and how lines and junctions are connected to each other, which limits its application in scene parsing and understanding.

\paragraph{Wireframe Parsing:} \cite{Huang:2018:LPW} proposes the wireframe parsing task. The authors train two separate neural networks to predict junction and line heat maps from an input image. After that, the two predictions are combined using a heuristic wireframe fusion algorithm to produce the final vectorized output. Although it is intuitive and can produce reasonable results, such two-stage process prevents the benefits of end-to-end training. In contrast, our framework is based on a single end-to-end trainable neural network, which directly delivers a vectorized wireframe representation as the output. 

\paragraph{Instance-level Recognition:} At the technical level, our method is inspired by instance-level recognition frameworks such as Fast R-CNN \cite{girshick2015fast}, Faster R-CNN \cite{ren2015faster}, CornerNet \cite{law2018cornernet}, Extremenet \cite{zhou2019bottom}. Our pipeline and LoI pooling (\Cref{sec:verification}) are conceptually similar to the RoI pooling in Faster R-CNN and Fast R-CNN. Both methods first generate a set of proposals and extract features to classify these proposes. The difference is that in \cite{ren2015faster, girshick2015fast}, the candidate proposals are generated by a sliding window fashion while our proposals are generated by connecting salient junctions (line sampler module \Cref{sec:proposal}). In this sense, the proposal generation procedure is also similar to what is used in point-based object detection \cite{law2018cornernet,zhou2019bottom}. The difference lies in how to discriminate between true lines and false positives. They use either similarity between points feature embedding \cite{law2018cornernet} or the classification score in the geometric center of several salient points \cite{zhou2019bottom} while ours extracts features to feed into a small neural network (line verification network \Cref{sec:verification}).

\section{Methods} \label{sec:method}

\begin{figure}[t]
    \centering
    \includegraphics[width=\linewidth]{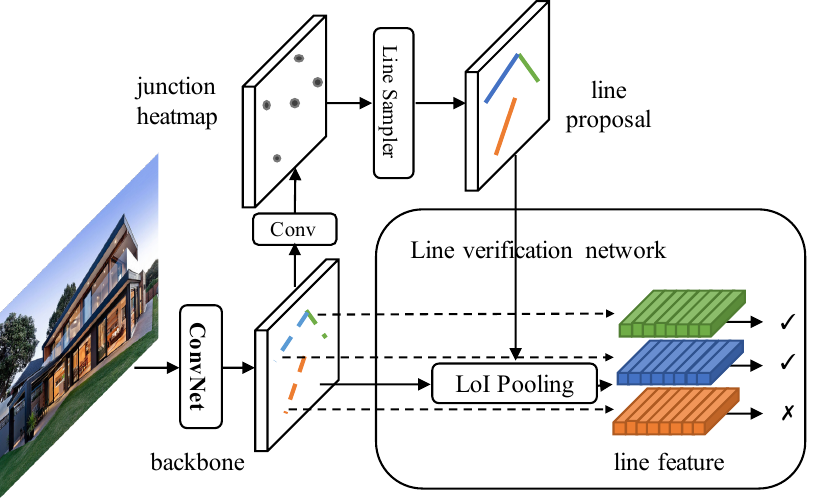}
    \caption{An overview of our network architecture.}
    \label{fig:pipeline}
\end{figure}

\subsection{Data Representation} 
\label{sec:representation}
Our representation of wireframes is based on the notation from graph theory. It can also be seen as a simplified version of the wireframe definition in \cite{Huang:2018:LPW}. Let $\wireframe=(\vertexSet,\edgeSet)$ be the wireframe of an image, in which the $\vertexSet$ is the set of junction indices and $\edgeSet \subseteq \vertexSet \times \vertexSet$ is the set of lines represented by the pair of junction endpoints in $\vertexSet$.  For each $i \in \vertexSet$, we use $\junction_i \in \mathbb{R}^2$ to represent the (ground truth) coordinate of the junction $i$ in the image space\requireJunctype{and $\junctionType_i$ be the type of that junction (if it is labeled in the dataset)}.

\subsection{Overall Network Architecture}
\label{sec:overallarch}

\Cref{fig:pipeline} illustrates the L-CNN architecture. It contains four modules: 1) a feature extraction backbone (\Cref{sec:backbone}) that takes a single image as the input and provides shared intermediate feature maps for the successive modules; 2) a junction proposal module (\Cref{sec:junction}) which outputs the candidate junctions; 3) a line sampling module (\Cref{sec:proposal}) that outputs line proposals based on the output junctions from the junction proposal module; 4) a line verification module (in \Cref{sec:verification}) which classifies the proposed lines. The output of L-CNN are the positions of junctions and the connectivity matrix among those junctions. Our system is fully end-to-end trainable with stochastic gradient descent.

\subsection{Backbone Network} \label{sec:backbone}
The function of the backbone network is to extract semantically meaningful features for the successive modules of L-CNN. We choose stacked hourglass network \cite{Newell:2016:Stacked} 
as our backbone for its efficiency and effectiveness. Input images are resized into squares. The stacked hourglass network first downsamples the input images twice in the spatial resolution via two 2-strided convolution layers. After that, learned feature maps are gradually refined by multiple U-Net-like modules \cite{ronneberger2015u} (the hourglass modules) with intermediate supervision imposed on the output of each module. The total loss of the network is the sum of the loss on those modules.

\subsection{Junction Proposal Module} \label{sec:junction}
\paragraph{Junction Prediction:}  We use a simplified version of \cite{Huang:2018:LPW} to estimate the candidate junction locations in the wireframe. An input image with resolution $W \times H$ is first divided into $W_b\times H_b$ bins. For each bin, the neural network predicts whether there exists a junction inside it, and if yes, it also predicts the its relative location inside this bin. Mathematically, the neural network outputs a junction likelihood map $\junctionMap$ and an offset map $\offsetMap$.  For each bin $\bin$, we have
\begin{equation*}
  \junctionMap(\bin) = 
  \begin{cases}
 1  & \exists i \in \vertexSet:  \junction_i \in b, \\
 0 & \text{otherwise} \\
  \end{cases} \\
\end{equation*}
and
\begin{equation*}
  \offsetMap(\bin) = \begin{cases}
  (\location - \junction_i)/W_b & \exists i \in \vertexSet:  \junction_i \in b,\\
 0 & \text{otherwise}. \\
  \end{cases}
\end{equation*}
where $\location$ represents the location of bin $\bin$'s center and $\junction$ represents the location of a vertex in $\vertexSet$.  \requireJunctype{For the datasets that have labels of junction types, we simply replace them with multiple junction maps $\junctionMap_t$ and offset maps $\offsetMap_t$, where $t$ represents the type of junctions.}

To predict $\junctionMap$ and $\offsetMap$, we design a network head that consists of two $1\times 1$ convolution layers to transform the feature maps into $\junctionMap$ and $\offsetMap$. We treat the problem of prediction $\junctionMap$ as a classification problem and use the average binary cross entropy loss. We use $\ell_2$ regression to predict the offset map $\offsetMap$. As the range of offset $\offsetMap(\bin)$ is bounded by $[-1/2, 1/2)\times[-1/2, 1/2)$, we append a sigmoid activation with offset $-0.5$ after the head to normalize the output. The loss on $\offsetMap$ is averaged over the bins that contain ground-truth junctions for each input image.

\paragraph{Non-Maximum Suppression:}

In instance-level recognition, non-maximum suppression (NMS) is applied to remove duplicate around correct predictions. We use the same mechanism for remove blurred score map around correct predictions and get $\junctionMap'(\bin)$ as:
\begin{equation*}
  \junctionMap'(\bin)  = \begin{cases}
    \junctionMap(\bin) & \junctionMap(\bin) = \max_{\bin'\in\mathcal{N}(\bin)} \junctionMap(\bin') \\
    0 & \text{otherwise}, \\
  \end{cases}
\end{equation*}
where $\mathcal{N}(\bin)$ represents the 8 nearby bins around $\bin$.  Here, we suppress the pixel values that are not the local maxima on the junction map. Such non-maximum suppression can be implemented with a max-pooling operator. The final output of the junction proposal network is the top $K$ junction positions $\{\hat \junction_i\}_{i=1}^K$ with the highest probabilities in $\junctionMap'$.

\subsection{Line Sampling Module} \label{sec:proposal}

\begin{figure}[t]
    \centering
    \begin{subfigure}[b]{0.32\linewidth}
        \centering
        \includegraphics[width=.99\linewidth]{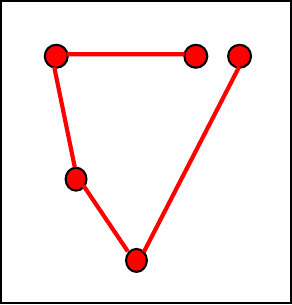} 
        \caption{Ground truth}
    \end{subfigure}
    \begin{subfigure}[b]{0.32\linewidth}
        \centering
        \includegraphics[width=.99\linewidth]{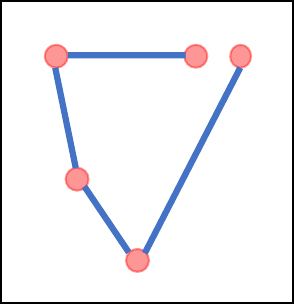} 
        \caption{Example of $\staticSet^+$}
    \end{subfigure}
    \begin{subfigure}[b]{0.32\linewidth}
        \centering
        \includegraphics[width=.99\linewidth]{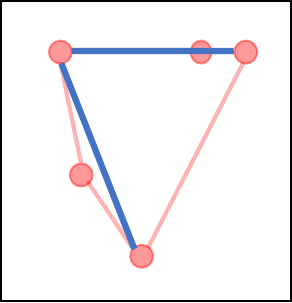} 
        \caption{Example of $\staticSet^-$}
    \end{subfigure}
    
    \vspace{5pt}
    
    \begin{subfigure}[b]{0.32\linewidth}
        \centering
        \includegraphics[width=.99\linewidth]{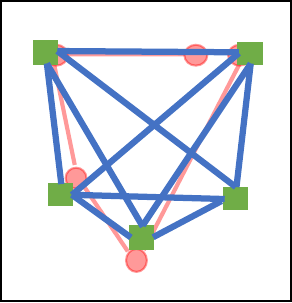} 
        \caption{Example of $\dynamicSet^*$}
    \end{subfigure}
    \begin{subfigure}[b]{0.32\linewidth}
        \centering
        \includegraphics[width=.99\linewidth]{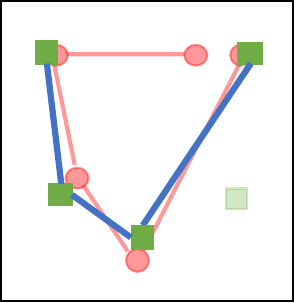} 
        \caption{Example of $\dynamicSet^+$}
    \end{subfigure}
    \begin{subfigure}[b]{0.32\linewidth}
        \centering
        \includegraphics[width=.99\linewidth]{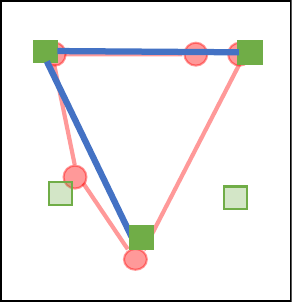} 
        \caption{Example of $\dynamicSet^-$}
    \end{subfigure}
    
    \caption{Illustration of our sampling methods. Red circles represent the ground truth junctions, red lines represent the ground truth lines, green squares represent the predicted junctions, and blue lines represent the candidate lines in the static and dynamic samplers.}
    \label{fig:sampler}
\end{figure}

Given a list of $K$ best candidate junctions $\{\hat \junction_i\}_{i=1}^K$\requireJunctype{ and potentially their junction types $\{\hat \junctionType_i\}_{i=1}^K$} from the junction proposal module, the purpose of the line sampling module is to generate a list of line candidates $\{L_j\}_{j=1}^M=\{(\tilde \junction^1_j, \tilde \junction^2_j\requireJunctype{, \tilde t_j^1, \tilde t_j^2})\}_{j=1}^M$ during the training stage so that the line verification network can learn to predict the existence of a line. Here $\tilde \junction_j^1$ and $\tilde \junction_j^2$ represents the coordinates of two endpoints of the $j$th candidate line segment. In this task, the amount of positive samples and negatives samples are extremely unbalanced, we address this issue by carefully design the sampling mechanism as stated below.

\paragraph{Static Line Sampler:}
For each image, the static line sampler returns  $N_\mathrm{\staticSet^+}$ positive samples and $N_\mathrm{\staticSet^-}$  negative samples that are directly derived from the ground truth labels. We call them static samples since they are irrelevant to the predicted candidate junction positions. 
Positive line samples are uniformly sampled from all the ground truth lines, denoted by $\staticSet^+$, with the ground truth coordinate of the corresponding junctions. 
The number of total negative line samples is $O(|\vertexSet|^2)$, which is huge compared to the number of positive samples $O(|\edgeSet|)$. To alleviate the problem, we sample the negative lines from $\staticSet^-$, a set of negative lines that are potentially hard to classify. We use the following heuristic to compute the $\staticSet^-$: we first rasterize all the ground truth lines onto a $64 \times 64$ low-resolution bitmap. Then, for each possible connections formed by a pair of ground truth junctions that is not a ground truth line, we define its \emph{hardness score} to be the average pixel density on the bitmap along this line. For each image, $\staticSet^-$ is set to be the top 2000 lines with the highest hardness scores.

\paragraph{Dynamic Line Sampler:}
In contrast to the static line sampler, the dynamic line sampler samples the lines using the predicted junctions from the junction proposal module. The sampler first matches all the predicted junctions to the ground truth junction. Let 
$  m_i:=\argmin_{j\requireJunctype{:\,\hat\junctionType_i = \junctionType_j.}} \|\hat \junction_i-\junction_j\|_2  $
be the index of the best matching ground truth junction for the $i$th junction candidates. If the $\ell_2$-distance between $\hat \junction_i$ and $\junction_{m_i}$ is less than the threshold $\eta$, we say that the junction candidate $\hat \junction_i$ is \emph{matched}. For each line candidate line $(\hat \junction_{i_1}, \hat \junction_{i_2}\requireJunctype{, \hat\junctionType_{i_1}, \hat\junctionType_{i_2}})$ in which $i_1, i_2 \in \{1,2,\dots, K\}$ and $i_1 \ne i_2$, we put it into line sets $\dynamicSet^+$, $\dynamicSet^-$, and $\dynamicSet^*$ according to the following criteria:
\begin{itemize}
    \item if both $\hat \junction_{i_1}$ and $\hat \junction_{i_2}$ are matched, and $(m_{i_1},m_{i_2}) \in \edgeSet$, we add this line to the \emph{positive sample set} $\dynamicSet^{+}$;
    \item if both $\hat \junction_{i_1}$ and $\hat \junction_{i_2}$ are matched, and $(m_{i_1},m_{i_2}) \in \staticSet^-$, we add this line to the \emph{hard negative sample set} $\dynamicSet^{-}$;
    \item the \emph{random sample set} $\dynamicSet^*$ includes all the line candidates from the predicted junctions, regardless of their matching results.
\end{itemize}
Finally, we randomly choose $N_\mathrm{\dynamicSet^{+}}$ lines from the positive sample set, $N_\mathrm{\dynamicSet^{-}}$ lines from the hard negative sample set, $N_\mathrm{\dynamicSet^*}$ lines from the random line sample set, and return their union as the dynamic line samples.

On one hand, the static line sampler helps cold-start the training at the beginning when there are few accurate positive samples from the dynamic sampler. It also complements the dynamic sampler by adding ground truth positive samples and hard negative samples to help the joint training process. On the other hand, the dynamic line sampler improves the performance of line detection by adapting the line endpoints to the predicted junction locations.

\subsection{Line Verification Network} \label{sec:verification}
The line verification network takes a list of candidate lines $\{L_j\}_{j=1}^M=\{(\tilde \junction^1_j, \tilde \junction^2_j\requireJunctype{, \tilde t_j^1, \tilde t_j^2})\}_{j=1}^M$ along with the feature maps of the image from the backbone network as the input and predicts whether or not each line is in the wireframe of the scene. During training, $L$ is computed by the line sampling modules, while during the evaluation, $L$ is set to be every pair of the predicted junctions $\{\hat \junction_i\}_{i=1}^K$.

For each candidate line segment $L_j=(\tilde \junction^1_j, \tilde \junction^2_j\requireJunctype{, \tilde t_j^1, \tilde t_j^2})$, we feed the coordinates of its two endpoints into a \emph{line of interest (LoI) pooling layer} (introduced below), which returns a fixed-length feature vector. Then, we pass the concatenated feature vector into a network head composed of two fully connected layers and get a logit. The loss of the line is the sigmoid binary cross entropy loss between the logit and the label of this line, i.e., a positive sample or a negative sample. To keep the loss balanced between positive and negative samples, the loss on each image for the line verification network is the sum of two separated loss, averaged over the positive lines and the negative lines, respectively.

\paragraph{LoI Pooling:} To check whether a line segment exists in an image, we first turn the line into a feature vector. Inspired by the RoIPool and RoIAlign layers from the object detection community \cite{girshick2014rich,girshick2015fast,ren2015faster,he2017mask}, we propose the LoI pooling layer to extract line features while it can back-propagate the gradient to the backbone network.

Each LoI is defined by the coordinates of its two endpoints, i.e.,  $\tilde\junction^1_j$ and $\tilde \junction^2_j$.  The LoI pooling layer first computes the coordinates of $N_p$ uniform spaced middle points along the line with linear interpretation
\begin{equation*}
    \locationA_k = \frac{k}{N_p-1}\tilde\junction^1_j + \frac{N_p-k}{N_p-1}\tilde\junction^2_j,\quad \forall k \in \{0, 1,\dots,N_p-1\}.
\end{equation*}
Then, it calculates the feature values at those $N_p$ points in the backbone's feature map using bilinear interpretation to avoid quantization artifacts \cite{dai2016instance, jaderberg2015spatial, dai2017deformable, he2017mask}. The resulting feature vector has a spatial extent of $C \times N_p$, in which $C$ is the channel dimension of the feature map from the backbone network.  After that, the LoI Pooling layer reduces the size of the feature vector with a 1D max pooling layer.  The result feature vector has shape $C \times \lceil \frac{N_p}{s} \rceil$, where $s$ is the size of stride of the max pooling layer.  This vector is then flattened and returned as the output of LoI pooling layer.

\section{Experiments} \label{sec:experiments}
\subsection{Implementation Details} \label{sec:implementation}

We use a stacked hourglass network \cite{Newell:2016:Stacked} as our backbone. Given an input image, we first apply a $7\times7$ stride-2 convolution, three residual blocks with channel dimension 64, and append a stride-2 max pooling on it. Then this intermediate feature representation is fed into two stacked hourglass modules. In each hourglass, the feature maps are down-sampled with 4 stride-2 residual blocks and then up-sampled with nearest neighbour interpolation. The dimensions of both the input channel and the output channel of each residual block are $256$. The network heads for $\junctionMap$ and $\offsetMap$ contain a $3\times3$ convolutional layer that reduces the number of channels to 128 with the ReLU non-linearity, followed by a $1\times1$ convolutional layer to match the output dimension. 

We reduce the feature dimension from 256 to 128 using a $1\times 1$ convolution kernels before feeding the feature map into the line verification network. For the LoIPool layer, we pick $N_p=32$ points along each line as the features of the line, resulting a $128 \times 32$ feature for each line. After that, we apply a one-dimensional stride-4 max pooling to reduce the spatial dimension of line features from 32 to 8. Our final line feature has dimension $128 \times 8$. The head of the line verification network then takes the flattened feature vector and feeds it into two fully connected layers with ReLU non-linearity, in which the middle layer has 1024 neurons.

All the experiments are conducted on a single NVIDIA GTX 1080Ti GPU for neural network training.  We use the ADAM optimizer \cite{kingma2014adam}. The learning rate and weight decay are set to $4 \times 10^{-4}$ and $1 \times 10^{-4}$, respectively. The batch size is set to $6$ for maximizing the GPU memory occupancy.  We train the network for 10 epochs and then decay the learning rate by 10.  We stop the training at 16 epochs as the validation loss no longer decreases. The total training time is about 8 hours. All the input images are resized to $(H,W)=(512, 512)$ and we use $H_b \times W_b=128 \times 128$ bins for $\junctionMap$ and $\offsetMap$.  
The junction proposal proposal network outputs the best $K=300$ junctions.  
For the line sampling module, we use $N_{\staticSet^+}=300$, $N_{\staticSet^-}=40$, $N_{\dynamicSet^+}=300$, $N_{\dynamicSet^-}=80$, and $N_{\dynamicSet^*}=600$.  The loss weights of multi-task learning for $\junctionMap$, $\offsetMap$, and line verification network are set to 8, 0.25, and 1, respectively.  Those weights are adjusted so that the magnitudes of the losses are in a similar scale.

\subsection{Datasets} \label{sec:dataset}
We conduct most of our experiments on the ShanghaiTech dataset \cite{Huang:2018:LPW}. It contains 5,462 images of man-made environments, in which 5000 images are used as the training set and 462 images are used as the testing set. The wireframe annotation of this dataset includes the positions of the salient junctions $\vertexSet$ and lines $\edgeSet$.  \N{We also test the models trained with the ShanghaiTech dataset on the York Urban dataset \cite{denis2008efficient} to evaluate the generalizability of all the methods.}

\subsection{Evaluation Metric} \label{sec:metric}

\begin{figure}[t]
    \centering
    \begin{subfigure}[b]{0.49\linewidth}
        \centering
        \includegraphics[width=0.49\linewidth]{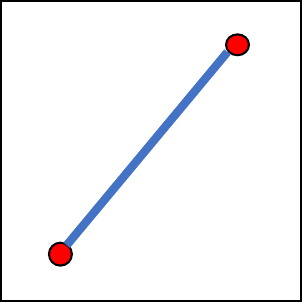} 
        \includegraphics[width=0.49\linewidth]{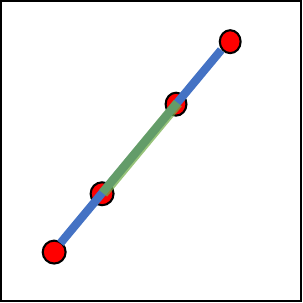}
        
        \includegraphics[width=0.49\linewidth]{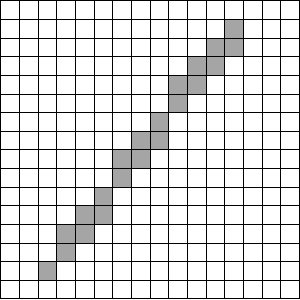}
        \includegraphics[width=0.49\linewidth]{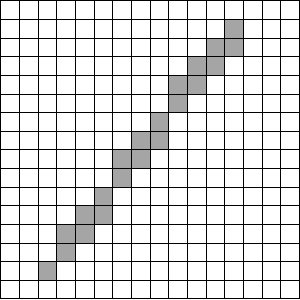}
        \caption{Overlapped lines}
        \label{fig:duplicated-line}
    \end{subfigure}
    \begin{subfigure}[b]{0.49\linewidth}
        \centering
        \includegraphics[width=0.49\linewidth]{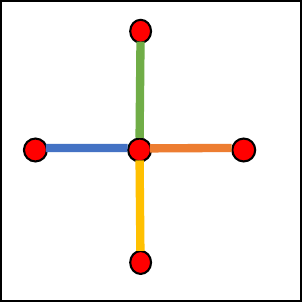} 
        \includegraphics[width=0.49\linewidth]{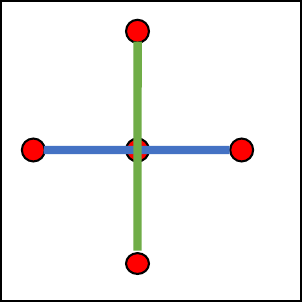}
        
        \includegraphics[width=0.49\linewidth]{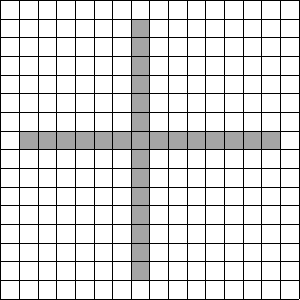}
        \includegraphics[width=0.49\linewidth]{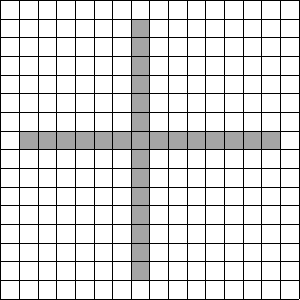}
        \caption{Incorrect connectivity}
        \label{fig:incorrect-connection}
    \end{subfigure}
    \caption{Demonstration of cases that heat map-based metric is not ideal for wireframe quality evaluation.  The upper part shows the detected lines and their heat maps, and the lower part shows the ground truth lines and their heat maps.}
    \label{fig:bad-metric}
\end{figure}

Previously, researchers use two metrics to evaluate the quality of detected wireframes: the heat map-based AP${}^{\mathrm{H}}$ to evaluate lines and junction AP to evaluate junctions. In this section, we first give a brief introduction to these metrics and discuss the reason why they are not proper for the wireframe parsing tasks. Then we \N{give} a new metric, named {\em structural AP}, a more reasonable way to evaluate the structural quality of wireframes.

\paragraph{Precision and Recall of Line Heat Maps:} The precision and recall curve over line heap maps is often used to evaluate the performance of wireframe and line detection \cite{Huang:2018:LPW, xue2018learning}.  Given a vectorized representation (lines or wireframes), it first generates a confidence heat map by rasterizing the lines. To compare it with the ground truth heat map, a bipartite matching that treats each pixel independently as a graph node is ran to match between two heat maps. Then precision and recall curve is computed according to the matching and confidence of each pixel.  In our experiment, we provide analysis of different methods using this metric.  We show both the F-score (as in \cite{xue2018learning}) and the area under the PR curve (similar to \cite{everingham2010pascal}) as the quantitative measure, and write the them as  $\text{F}^{\mathrm{H}}$ and $\text{AP}^{\mathrm{H}}$, respectively.

These metrics were originally designed for evaluating boundary detection \cite{martin2004learning} and they work well for that purpose.  However they are problematic in wireframe detection since
\begin{enumerate}[nosep]
    \item they do not penalize for overlapped lines (\Cref{fig:duplicated-line});
    \item they do not properly evaluate the connectivity of the wireframe (\Cref{fig:incorrect-connection}).  
\end{enumerate}
For example, if a long line is broken into several short line segments, the resulted heat map is almost the same as the ground truth heat map, as shown in \Cref{fig:bad-metric}. A good performance on the above two properties is vital for downstream tasks that rely on the correctness of line connectivity, such as inferring the 3D geometry through lines \cite{ramalingam2013lifting,zhou2019learning}.

\paragraph{Structural AP:} To overcome those drawbacks, we propose a new evaluation metric defined on vectorized wireframes rather than on a heat map.  We call our metric \emph{structural average precision (sAP)}.  This metric is inspired by the mean average precision commonly used in object detection \cite{everingham2010pascal}.  Structural AP is defined to be the area under the precision recall curve computed from a scored list of the detected line segments on all test images.  Recall is the proportion of the correctly detected line segments (up to a cutoff score) to all the ground truth line segments, while precision is the proportion of the correctly detected line segments above that cutoff to all the detected line segments.

A detected line segment $L_j=(\tilde\junction_j^1, \tilde\junction_j^2)$ is considered to be a true positive (correct) if and only if
\begin{equation*}
    \min_{(u,v) \in \edgeSet} \|\tilde\junction_j^1-\junction_u\|_2^2 + \|\tilde\junction_j^2-\junction_v\|_2^2 \le \threshold,
\end{equation*}
where $\threshold$ is a user-defined number represents the strictness of the metric.  In this experiment section, we evaluate the structural AP at $\threshold=5$, $\threshold=10$, and $\threshold=15$ under the resolution of $128 \times 128$.  We abbreviate them as sAP${^5}$, sAP${}^{10}$, and sAP${}^{15}$, respectively.  In addition, each ground truth line segment is not allowed to be matched more than once in order to penalize double-predicted lines. That is to say if there exists a line $L_i$ that is ranked above the line $L_j$ and
\begin{align*}
&\argmin_{(u,v) \in \edgeSet} \|\tilde\junction_i^1-\junction_u\|_2^2 + \|\tilde\junction_i^2-\junction_v\|_2^2 \\ ={}&\argmin_{(u,v) \in \edgeSet} \|\tilde\junction_j^1-\junction_u\|_2^2 + \|\tilde\junction_j^2-\junction_v\|_2^2,
\end{align*}
then the line $L_j$ will always be marked as a false positive.

\paragraph{Junction mAP:}  The major difference between line detection and wireframe detection is that the wireframe representation encodes junction positions. Junctions have physical meaning in 3D (corners or occlusional points) and encodes the line connectivity information. Our junction mean AP ($\text{mAP}^{\mathrm{J}}$) evaluates the quality of vectorized junctions of a wireframe detection algorithm without relying on heat maps as in \cite{Huang:2018:LPW}. To better understand the advantage of explicitly modeling junctions, we also evaluate our method using the junction mAP as described below: for a given ranked list of predicted junction positions, a junction is considered to be correct if the $\ell^2$ distance between this junction and its nearest ground-truth is within a threshold.  Each ground truth junction is only allowed to be matched once to penalize double-predicted junctions.  Using this criteria, we can draw the precision recall curve by counting the number of true and false positives.  The junction AP is defined to be the area under this curve.  The mean junction AP is defined to be the average of junction AP over difference distance thresholds. In our implementation, we choose to average over 0.5, 1.0, and 2.0 thresholds under $128 \times 128$ resolution.

\subsection{Ablation Study}

\begin{table}[t]
    \centering
    \small
    \setlength{\tabcolsep}{2.8pt}
    \renewcommand{\arraystretch}{1.1}
    \begin{tabular}{c|ccccc|cc|ccc}
    & \multicolumn{5}{c|}{sampler} & \multicolumn{2}{c|}{head} & \multicolumn{3}{c}{metric} \\
    & ${\dynamicSet}^{*}$ & ${\staticSet}^{+}$ & ${\staticSet}^{-}$ & ${\dynamicSet}^+$ & ${\dynamicSet}^{-}$ & fc+fc & conv+fc & sAP${}^{5}$ & sAP${}^{10}$ & sAP${}^{15}$ \\
    \hline
    \hline
    (a) & \checkmark & & & & & \checkmark & & 43.7 & 48.2 & 50.2 \\
    (b) & & \checkmark & \checkmark & & & \checkmark & & 38.5 & 41.9 & 43.8 \\
    (c) & \checkmark & \checkmark & \checkmark & & & \checkmark & & 47.8 & 51.7 & 53.6 \\
    (d) & \checkmark & \checkmark & \checkmark & \checkmark & \checkmark &  &\checkmark & 55.7 & 59.8 & 61.7 \\
    (e) & \checkmark & \checkmark & & \checkmark & & \checkmark & &   57.4 & 61.4 & 63.2 \\
    (f) & \checkmark & \checkmark & \checkmark & \checkmark & \checkmark & \checkmark & & \textbf{58.9} & \textbf{62.9} & \textbf{64.7} \\
    \end{tabular}
    \caption{Ablation study of L-CNN.  The columns labeled with ``sampler'' represent whether a specific sampler is used during the training stage, as introduced in \Cref{sec:proposal}.  The columns labeled with ``head'' represent the network head structured used in the line verification network.  ``fc + fc'' is the network structure introduced in \Cref{sec:verification}, while in ``conv + fc'' we replace the middle fully connected layer with a 1D Bottleneck layer \cite{he2016deep}.
    }
    \label{tab:ablation}
\end{table}

In this section, we run a series of ablation experiments on the ShanghaiTech dataset \cite{Huang:2018:LPW} to study our proposed method. We use our structural average precision (sAP) as the evaluation metrics. The results are shown in \Cref{tab:ablation}.

\paragraph{Line Sampling Modules:} We compare different design choices for line sampling modules, as shown in \Cref{tab:ablation}. (a) uses just the random pairs from the dynamic sampler. The sAP${}^5$ is 43.7, which serves as the baseline. (b) only uses the sampled pairs from ground-truth junctions and get much worse performance.  The performance gap is even larger when the evaluation criterion is loose. This is because (b) does not consider the imperfect of junction prediction map and cannot tackle when junction is slightly misaligned with the ground truth. After that, we combine the random dynamic sampling and the static sampling, which significantly improves the performance, as shown in \Cref{tab:ablation} (c). Then we add dynamic sampler candidate ${\dynamicSet}^+$ and ${\dynamicSet}^{-}$, which leads to the best sAP${}^5$ score 58.9 in (f). This experiment indicates that the carefully selected dynamic line candidates are vital to a good performance.  Lastly, by comparing (e) and (f), we find that including hard examples ${\staticSet}^-$ and ${\dynamicSet}^-$ is indeed helpful compared to just doing the random sampling in $\dynamicSet^*$.

\paragraph{Line Verification Networks:} \Cref{tab:ablation} also shows our ablation on how to design the line verification network. We tried two different designs: In \Cref{tab:ablation} (e), we apply two fully-connected layers after the LoI Pooling feature to get the classification results, while in \Cref{tab:ablation} (d) we firstly apply a 1D convolution on the features and then use the fully-connected layer on the flattened feature vector to get the final line classification. Experiments show that using convolution largely deteriorates the performance. We hypothesis that this is because line classification requires location sensitivity, which the translation-invariant convolution cannot provide \cite{dai2016r, li2017fully}.

\subsection{Comparison with Other Methods} \label{sec:comparision}

\begin{figure*}[t]
    \begin{subfigure}[b]{0.245\linewidth}
    \includegraphics[width=0.99\linewidth]{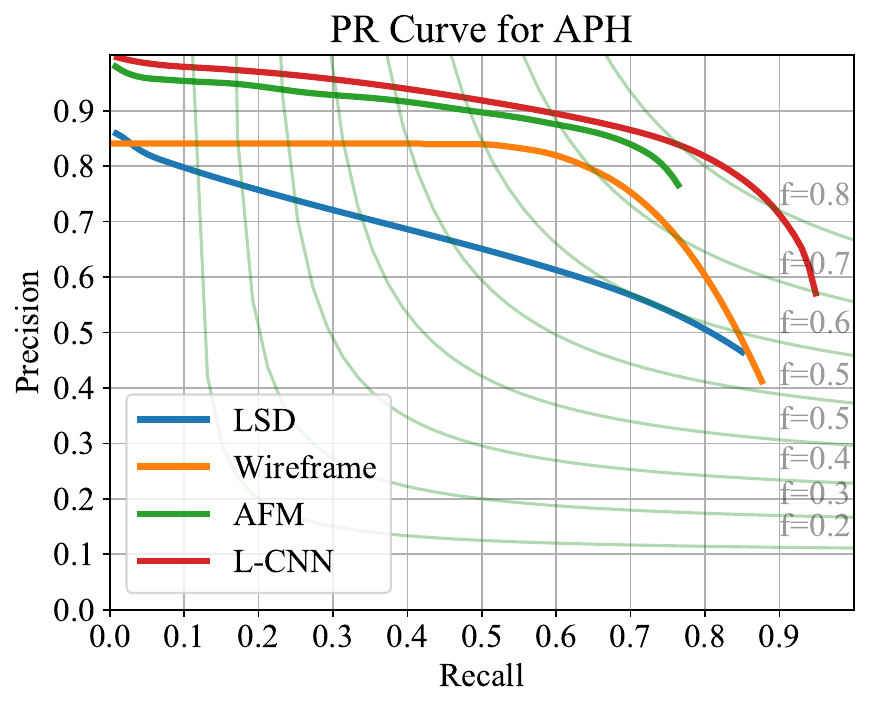}
    \caption{Heat map-based PR curves on the ShanghaiTech dataset \cite{Huang:2018:LPW}.}\label{fig:result:ap}
    \end{subfigure}
    \begin{subfigure}[b]{0.245\linewidth}
    \includegraphics[width=0.99\linewidth]{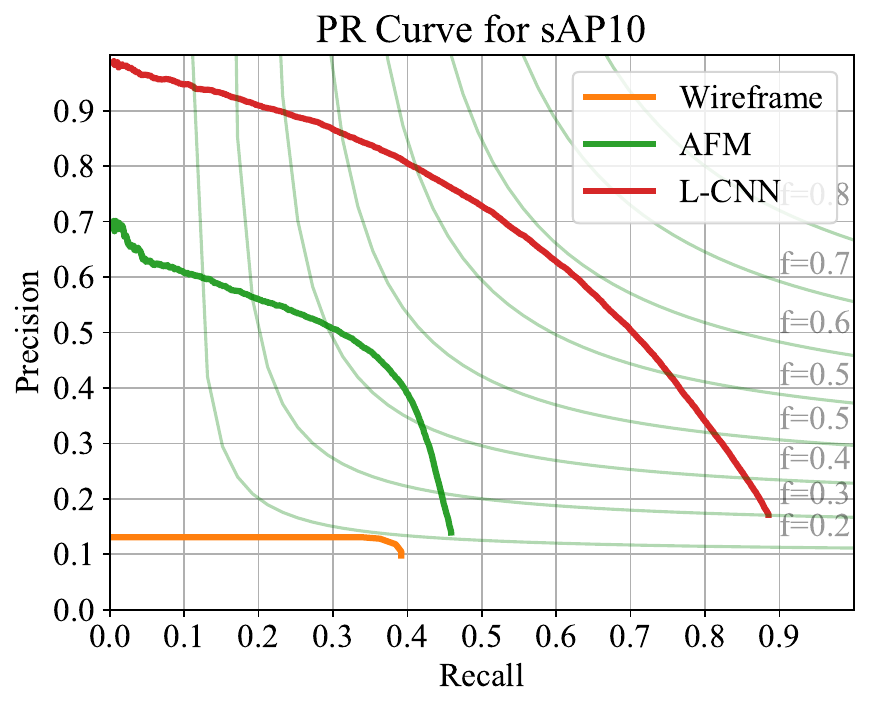}
    \caption{Structural PR curves on the ShanghaiTech dataset \cite{Huang:2018:LPW}.}\label{fig:result:sap}
    \end{subfigure}
    \begin{subfigure}[b]{0.245\linewidth}
    \includegraphics[width=0.99\linewidth]{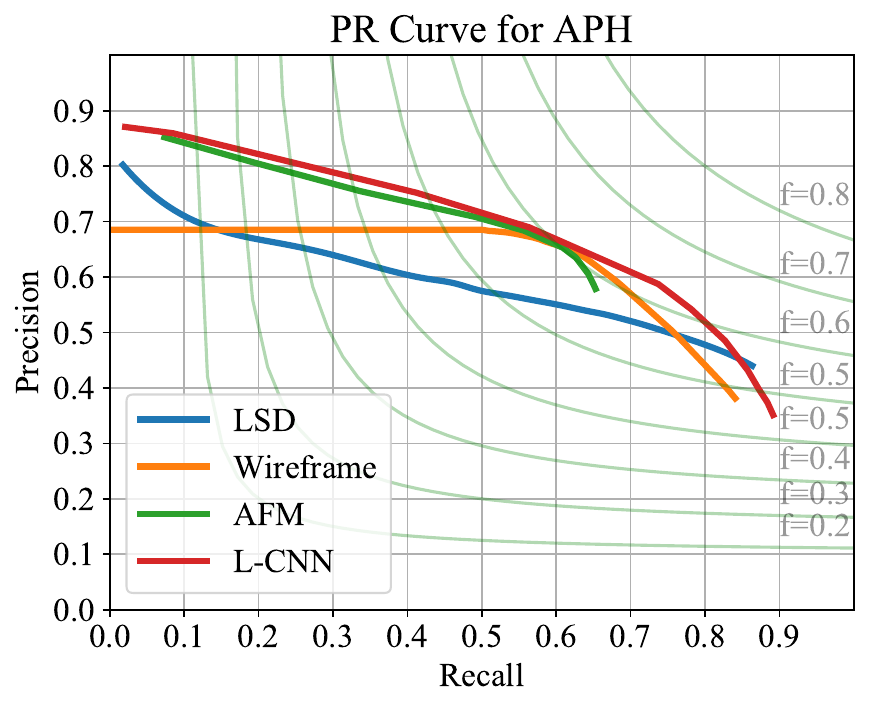}
    \caption{Heat map-based PR curves on the York Urban dataset \cite{denis2008efficient}.}\label{fig:result:ap2}
    \end{subfigure}
    \begin{subfigure}[b]{0.245\linewidth}
    \includegraphics[width=0.99\linewidth]{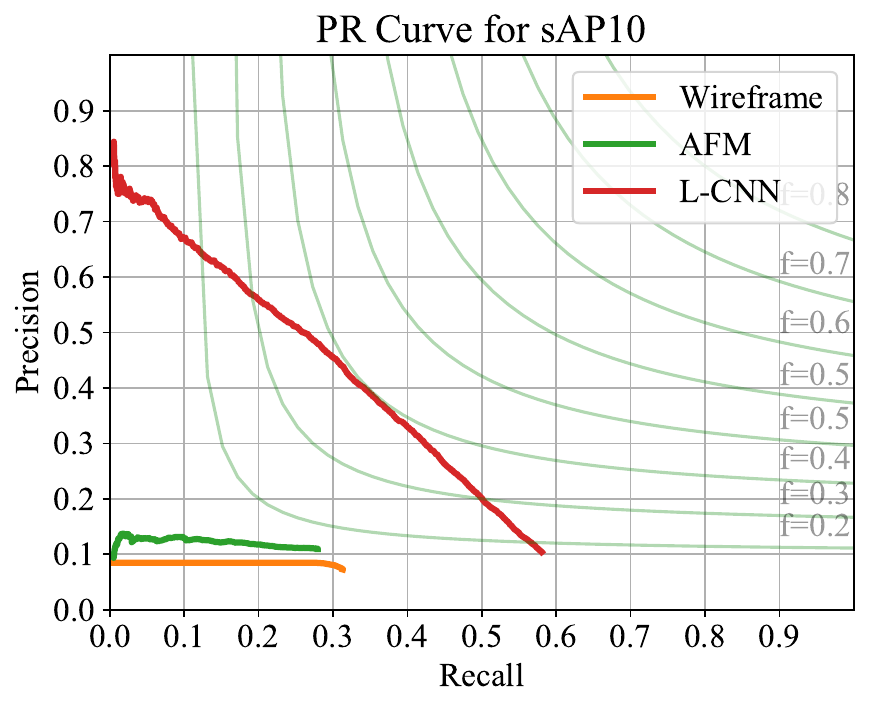}
    \caption{Structural PR curves on the York Urban dataset \cite{denis2008efficient}.}\label{fig:result:sap2}
    \end{subfigure}
    \caption{Precision recall curves of multiple algorithms.  Models are trained on ShanghaiTech .}\label{fig:result}
\end{figure*}

Following the practice of \cite{Huang:2018:LPW, xue2018learning}, we compare our method with LSD \cite{von2010lsd}, deep learning-based line detectors \cite{xue2018learning}, as well as wireframe parser from the ShanghaiTech dataset paper \cite{Huang:2018:LPW}. $\text{F}^{\mathrm{H}}$, $\text{AP}^{\mathrm{H}}$, and sAP with different thresholds are used to evaluate those methods quantitatively.  \N{All the models are trained on the ShanghaiTech dataset and evaluate on both of the ShanghaiTech \cite{Huang:2018:LPW} and York Urban datasets \cite{denis2008efficient}. The results are shown in \Cref{tab:result} and \Cref{fig:result}.}  We note that the difference of the numbers and curves for $\text{AP}^{\mathrm{H}}$ from \cite{Huang:2018:LPW, xue2018learning} is due to our more proper implementation of AP${}^{\mathrm{H}}$: 1) In the code provided by \cite{Huang:2018:LPW}, they evaluate the precision and recall per image and average them together, while we first sum the number of true positives and false positives over the dataset and then compute the precision and recall. 2) Due to the insufficient number of thresholds, the PR curves in \cite{Huang:2018:LPW, xue2018learning} do not cover all the recall that an algorithm can achieve. We evaluate all the methods on more thresholds to extend the curves as long as possible.

\Cref{fig:result:ap} shows that our algorithm is better than the state-of-the-art line detector methods under the PR curve of heat map-based line metrics, especially in the high-recall region. This indicates that our method can find more correct lines compared to other methods. We also quantitatively calculate the F-score and the average AP.   \Cref{tab:result} shows that our algorithm performs significantly better than previous state-of-the-art line detectors by 13.3 points in $\text{AP}^{\mathrm{H}}$ and 4.0 points in $\text{F}^{\mathrm{H}}$ \cite{xue2018learning}.  We also want to emphasize that compared to line detection, it is conceptually harder for the wireframe detection methods to reach the same performance as the line detection methods in term of the heat map-based metrics.  This is because a wireframe detection algorithm requires the positions of junctions, the endpoints of lines, to be correct, while a line detector can start and end a line arbitrarily to ``fill'' the line heat map.  Before evaluating the heat map-based metrics, we post process the lines from L-CNN to remove the overlap, as described in \N{Appendix A.1}.

We then evaluate all the methods with our proposed structural AP. The precision recall curve is shown in \Cref{fig:result:sap} (LSD is missing here as its scores are too low to be drawn). The gap between our method and previous methods is even larger.  Our method achieves 40-point sAP improvement over the previous state-of-the-art method.  This is because our line verification network penalizes incorrect structures, while methods such as AFM and wireframe parser use a hand-craft algorithm to extract lines from heat maps, in which the information of junction connectivity gets lost.  Furthermore, the authors of \cite{Huang:2018:LPW} mention that their vectorization algorithm will break lines and add junctions to better fit the predicted heat map.  Such behaviors can worsen the structure correctness, which might explain its low sAP score.%

The {$\text{mAP}^{\mathrm{J}}$} results are shown in \Cref{tab:result}. For AFM, we treat the endpoints of each line as junctions and use the line NFA score as the score of its endpoints. We note that the inferior junction quality of AFM is not because their method is not well-designed but the end task is different. This shows that one cannot directly apply a line detection algorithm on the wireframe parsing task. In addition, our L-CNN outperforms the previous wireframe parser \cite{Huang:2018:LPW} by a large margin due to the \N{joint training process of the pipeline}.

\Cref{tab:result} and \N{\Cref{fig:result:ap2,fig:result:sap2} show that L-CNN also performs the best among all the wireframe and line detection methods when testing on a different dataset \cite{denis2008efficient} without finetune.  This indicates that our method is able to generalize to novel scenes and data.  We note that the relatively low sAP scores are due to the duplicated lines, texture lines, while missing many long lines in the annotation of the dataset.}

\begin{table}[t]
    \centering
    \footnotesize
    \setlength{\tabcolsep}{3pt}
    \renewcommand{\arraystretch}{1.2}
\begin{tabular}{c|cc|cc|cc|cc}
&\multicolumn{4}{c|}{ShanghaiTech \cite{Huang:2018:LPW}}&\multicolumn{4}{c}{York Urban \cite{denis2008efficient}} \\
 & sAP${}^{10}$ & $\text{mAP}^{\mathrm{J}}$ & $\text{AP}^{\mathrm{H}}$ & $\text{F}^{\mathrm{H}}$ & sAP${}^{10}$ & $\text{mAP}^{\mathrm{J}}$ & $\text{AP}^{\mathrm{H}}$ & $\text{F}^{\mathrm{H}}$\\ \hline
LSD  & / & / & 52.0 & 61.0             & / & / & 51.0 & 60.0 \\ \hline
Wireframe & 9.0 & 36.1 & 67.8 & 72.6   & 3.9 & 22.5 & 53.4 & 63.7 \\ \hline
AFM  & 24.4 & 23.3 & 69.5 & 77.2       & 3.5 & 6.9 & 48.4 & 63.1\\ \hline
\textbf{L-CNN} & \textbf{62.9} & \textbf{59.3} & \textbf{82.8} & \textbf{81.2} & \textbf{26.4} & \textbf{30.4} & \textbf{59.8} & \textbf{65.4}
\end{tabular}

    \caption{Performance comparison of wireframe detection.  All the models are trained on the ShanghaiTech dataset and evaluate on both datasets.  The columns labelled with ``sAP'' show the line accuracy with respect to our structural metrics; the columns labelled with ``$\text{mAP}^{\mathrm{J}}$'' shows the mean average precision of the predicted junctions; the columns labelled with ``$\text{F}^{\mathrm{H}}$'' and ``$\text{AP}^{\mathrm{H}}$'' shows the performance metrics related to heat map-based PR curves.  %
    Our method L-CNN has the state-of-the-art performance on all of the evaluation metrics.}
    \label{tab:result}
\end{table}

\subsection{Visualization} \label{sec:visualization}

\begin{figure*}
    \centering\hfill
    \begin{minipage}[t]{0.19\linewidth}\centering
    \includegraphics[width=0.99\linewidth]{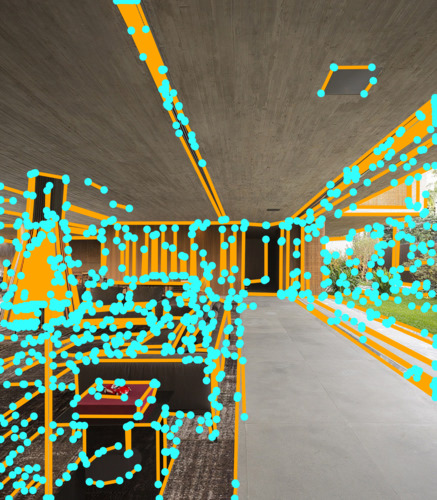}
    \includegraphics[width=0.99\linewidth]{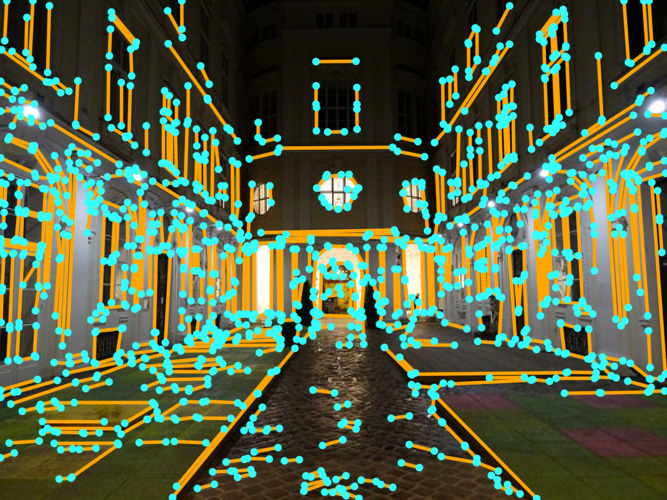}
    \includegraphics[width=0.99\linewidth]{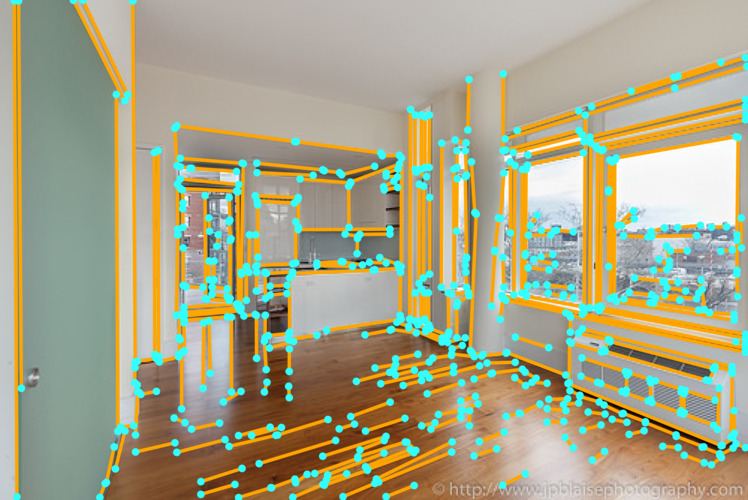}
    \includegraphics[width=0.99\linewidth]{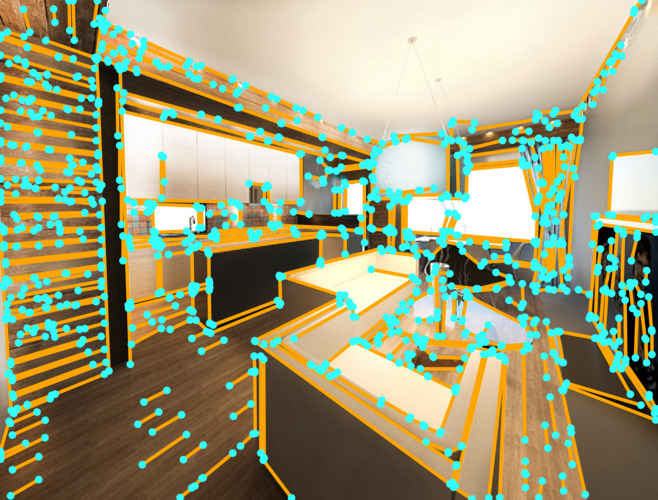}
    \includegraphics[width=0.99\linewidth]{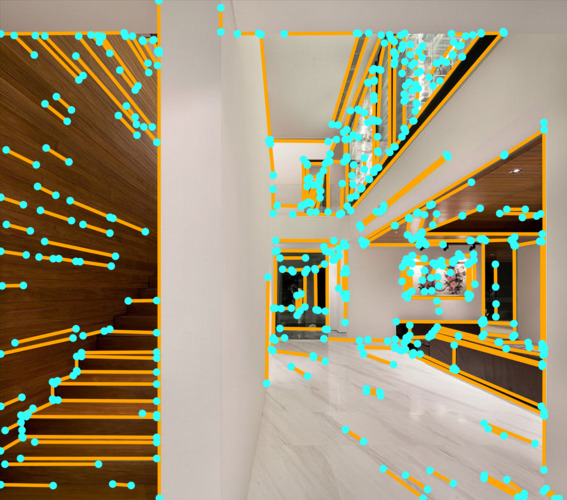}
    \includegraphics[width=0.99\linewidth]{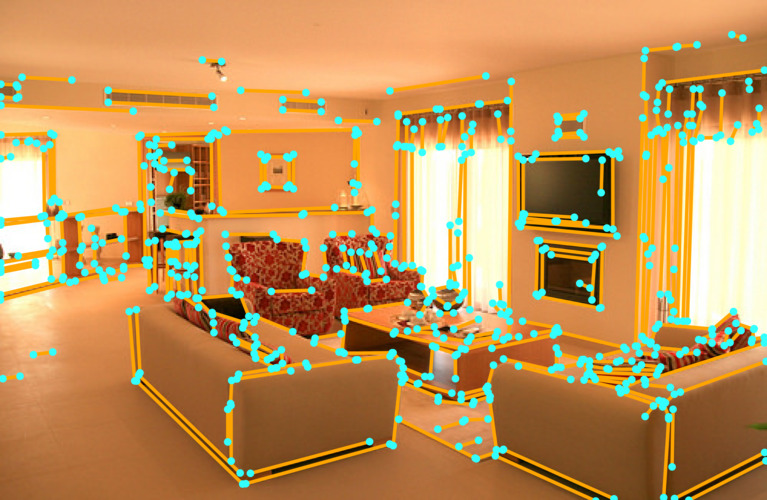}
    \includegraphics[width=0.99\linewidth]{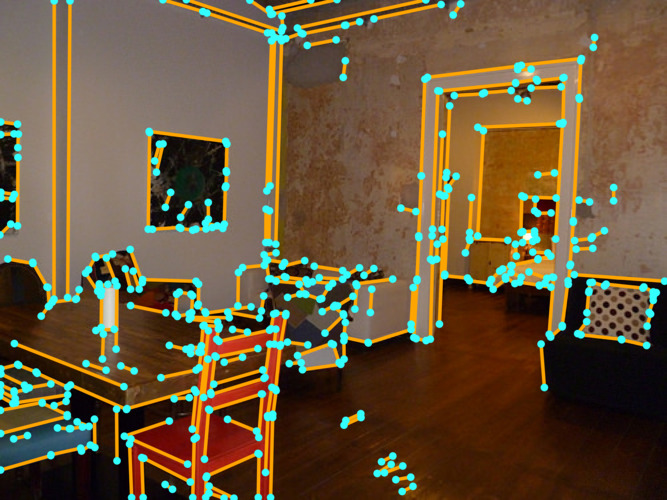}
    \includegraphics[width=0.99\linewidth]{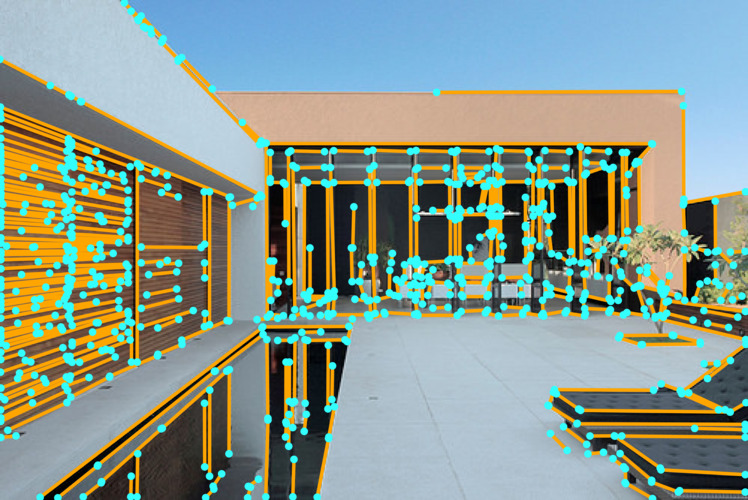}
    
    (a) LSD
    \end{minipage}\hfill
    \begin{minipage}[t]{0.19\linewidth}\centering
    \includegraphics[width=0.99\linewidth]{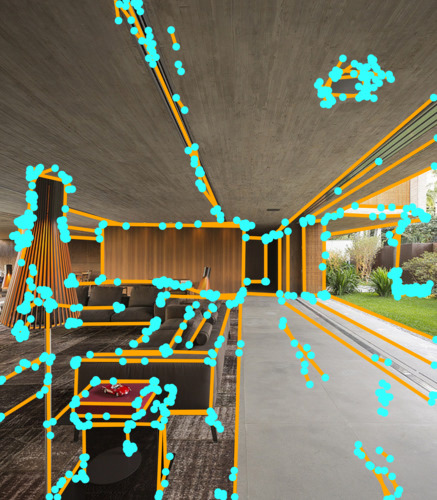}
    \includegraphics[width=0.99\linewidth]{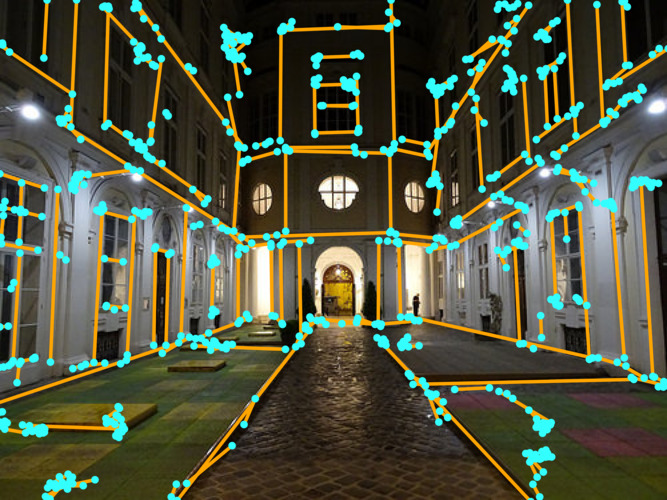}
    \includegraphics[width=0.99\linewidth]{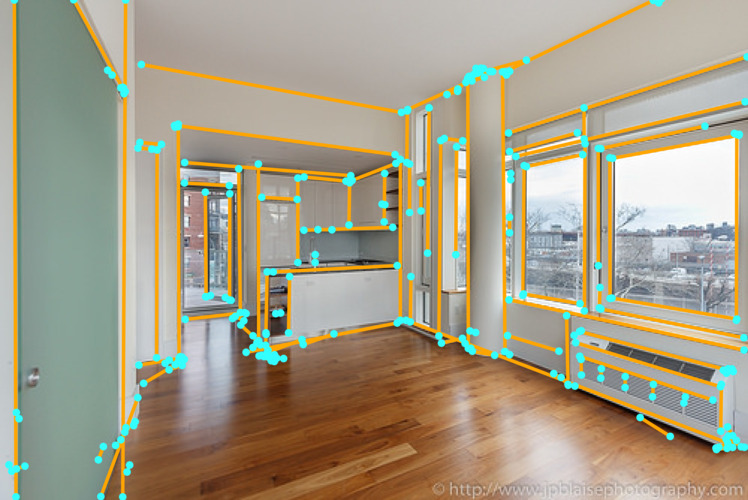}
    \includegraphics[width=0.99\linewidth]{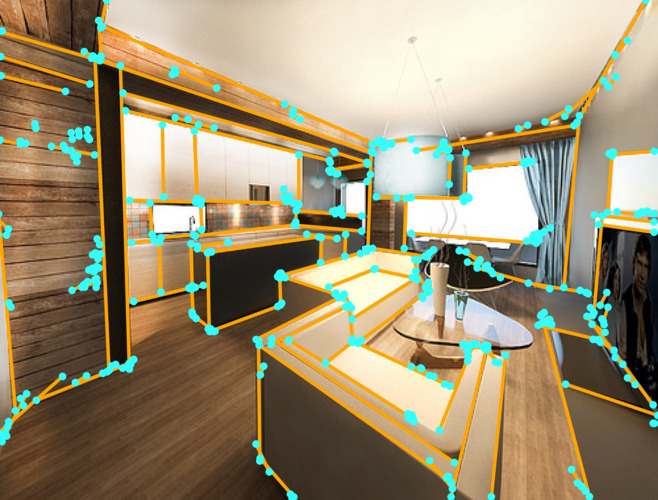}
    \includegraphics[width=0.99\linewidth]{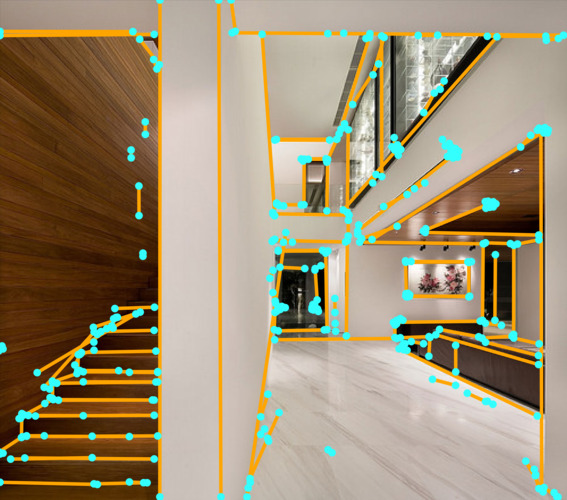}
    \includegraphics[width=0.99\linewidth]{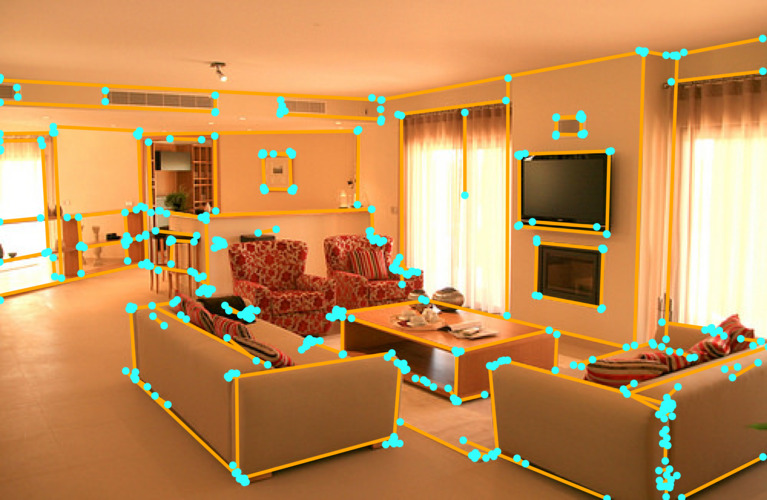}
    \includegraphics[width=0.99\linewidth]{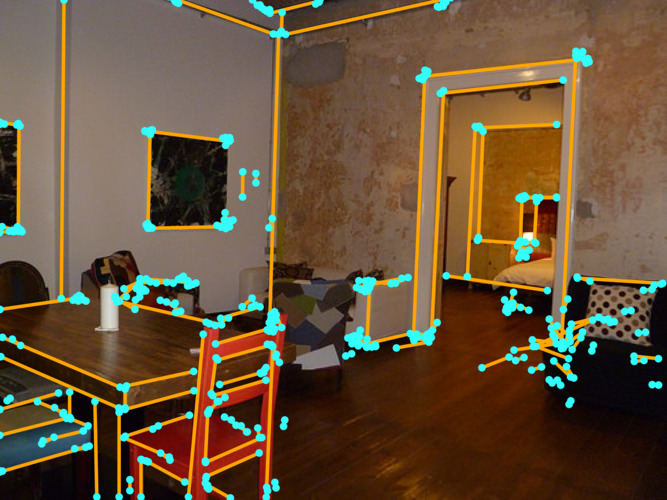}
    \includegraphics[width=0.99\linewidth]{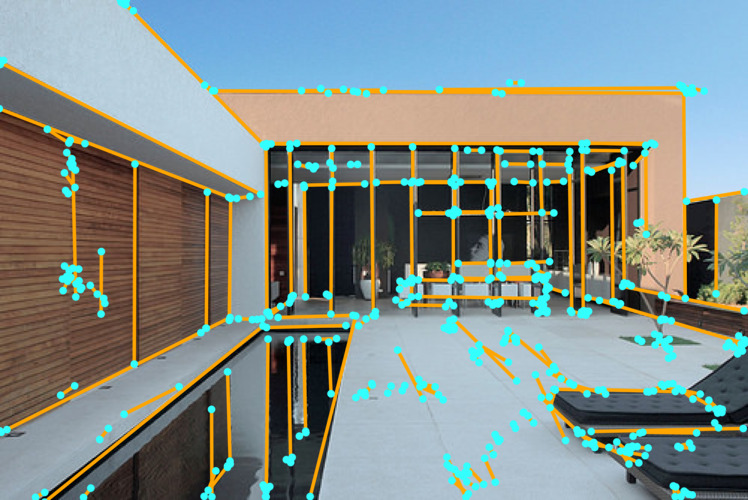}

    (b) AFM
    \end{minipage}\hfill
    \begin{minipage}[t]{0.19\linewidth}\centering
    \includegraphics[width=0.99\linewidth]{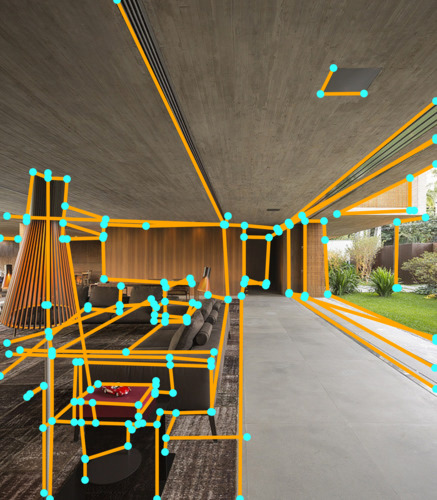}
    \includegraphics[width=0.99\linewidth]{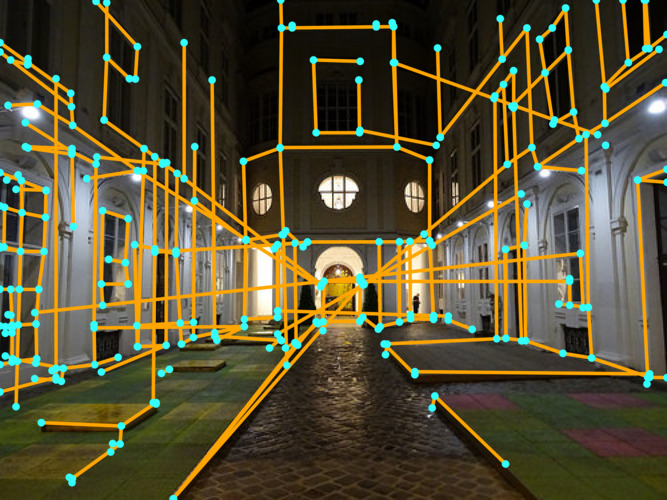}
    \includegraphics[width=0.99\linewidth]{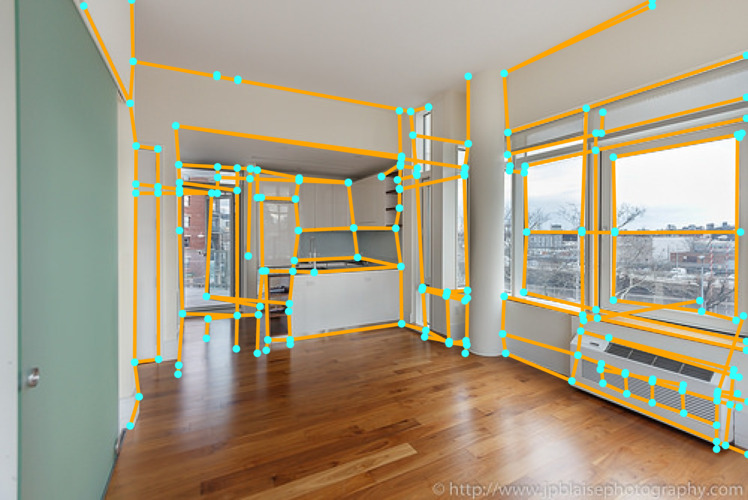}
    \includegraphics[width=0.99\linewidth]{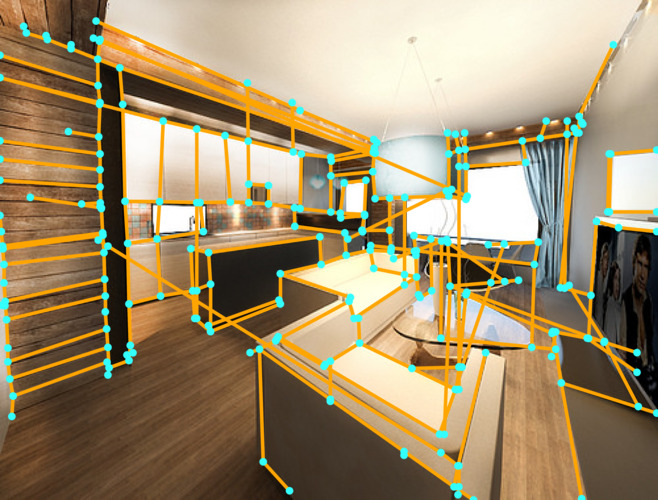}
    \includegraphics[width=0.99\linewidth]{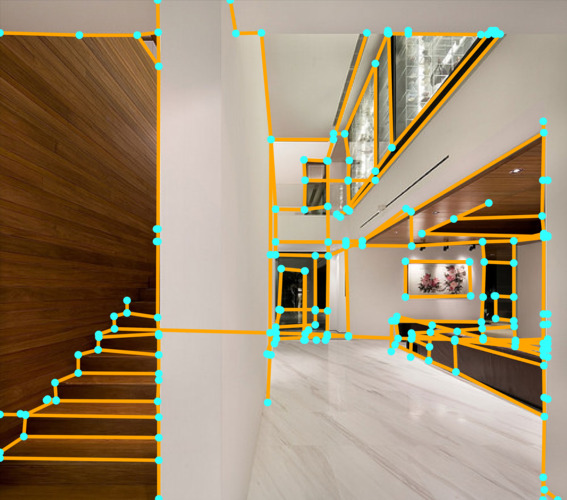}
    \includegraphics[width=0.99\linewidth]{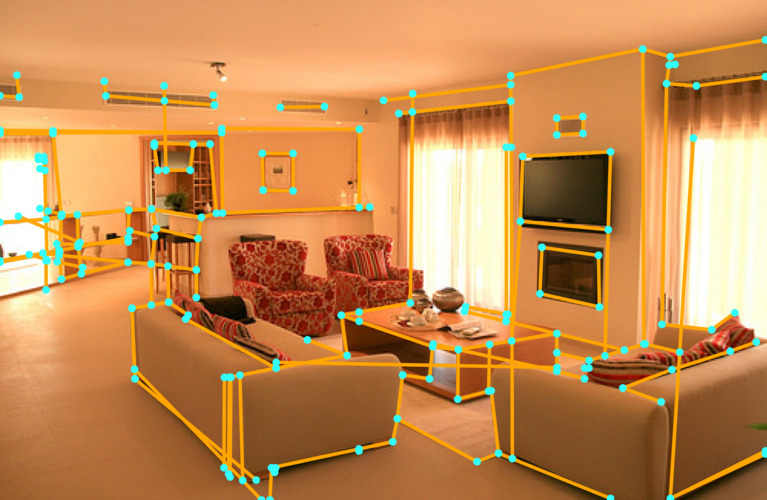}
    \includegraphics[width=0.99\linewidth]{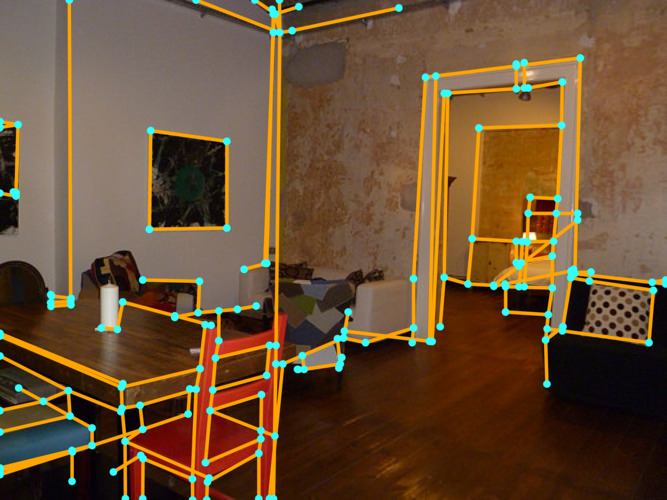}
    \includegraphics[width=0.99\linewidth]{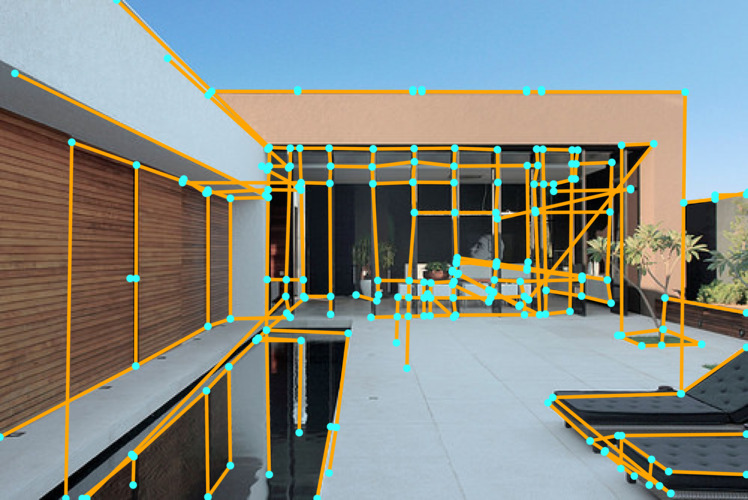}
    
    (c) Wireframe
    \end{minipage}\hfill
    \begin{minipage}[t]{0.19\linewidth}\centering
    \includegraphics[width=0.99\linewidth]{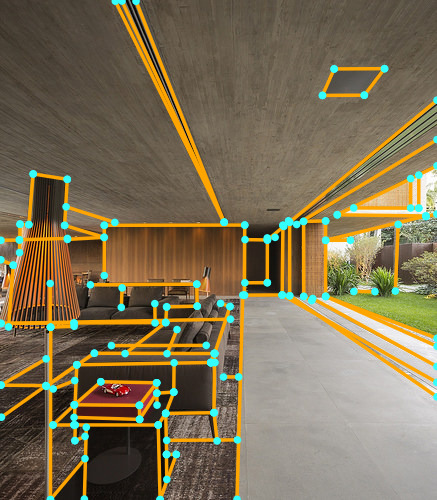}
    \includegraphics[width=0.99\linewidth]{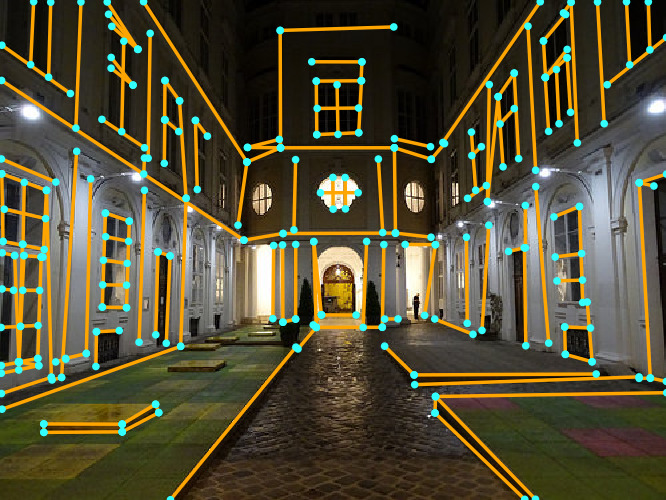}
    \includegraphics[width=0.99\linewidth]{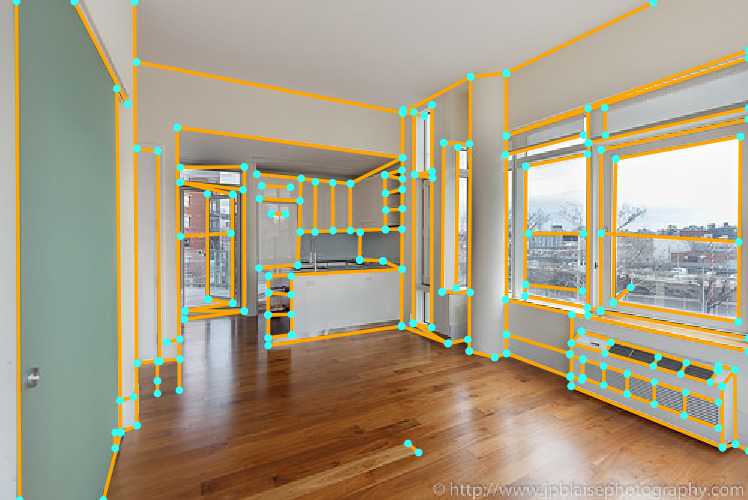}
    \includegraphics[width=0.99\linewidth]{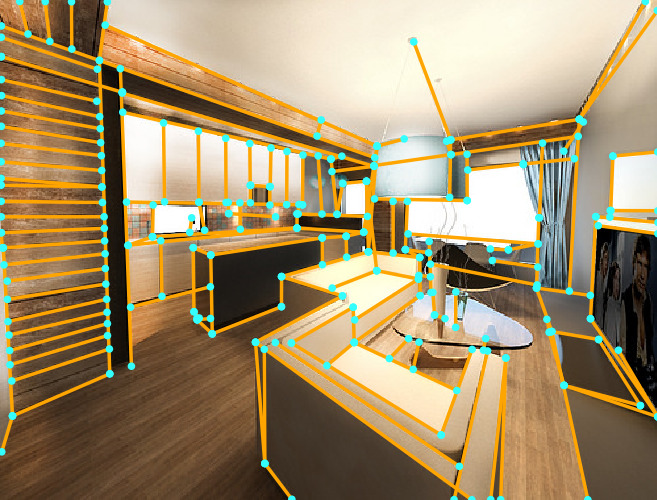}
    \includegraphics[width=0.99\linewidth]{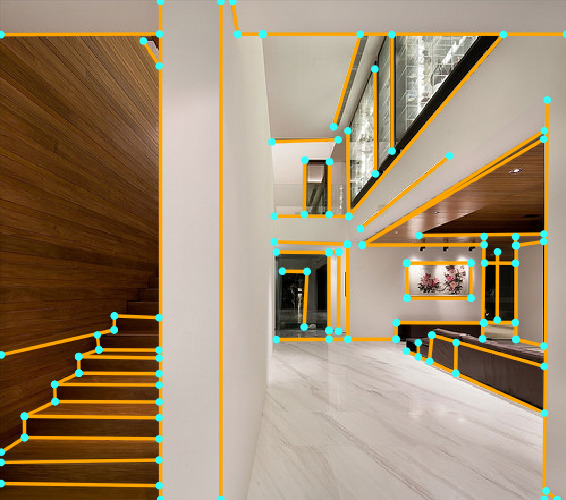}
    \includegraphics[width=0.99\linewidth]{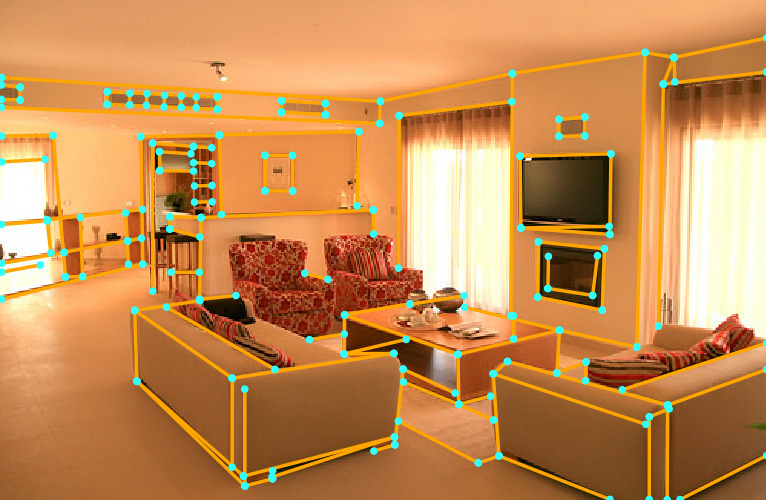}
    \includegraphics[width=0.99\linewidth]{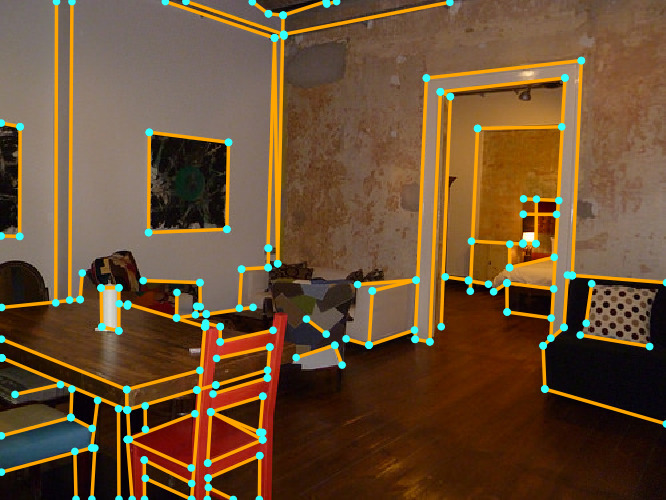}
    \includegraphics[width=0.99\linewidth]{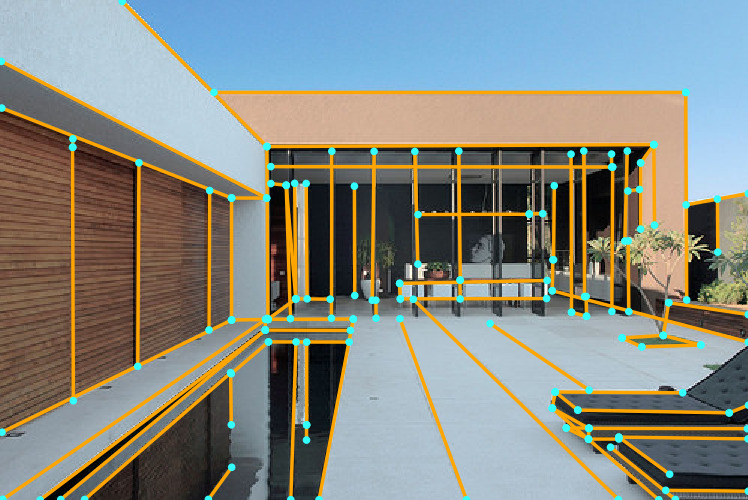}
    
    (d) L-CNN (ours)
    \end{minipage}\hfill
    \begin{minipage}[t]{0.19\linewidth}\centering
    \includegraphics[width=0.99\linewidth]{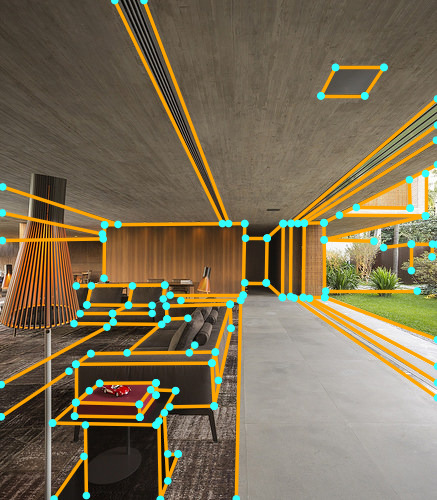}
    \includegraphics[width=0.99\linewidth]{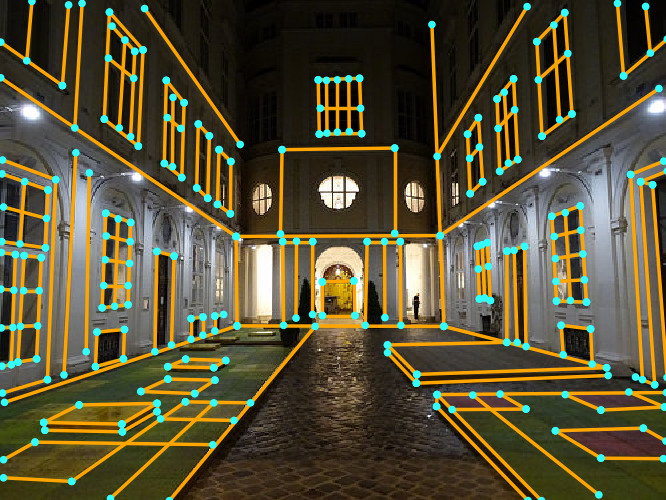}
    \includegraphics[width=0.99\linewidth]{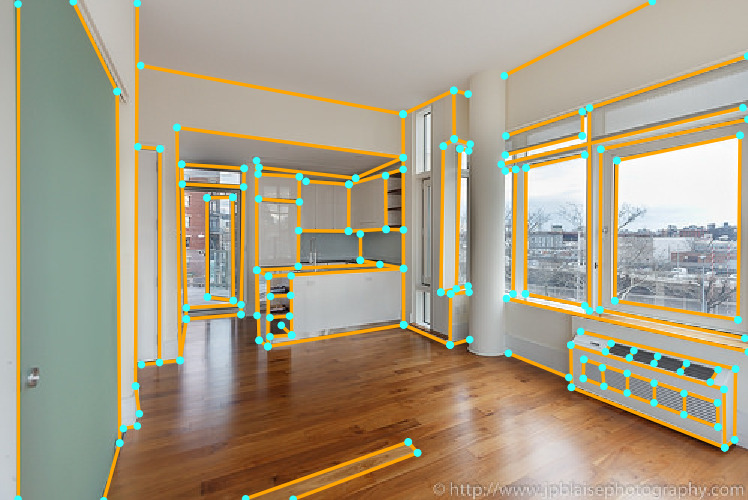}
    \includegraphics[width=0.99\linewidth]{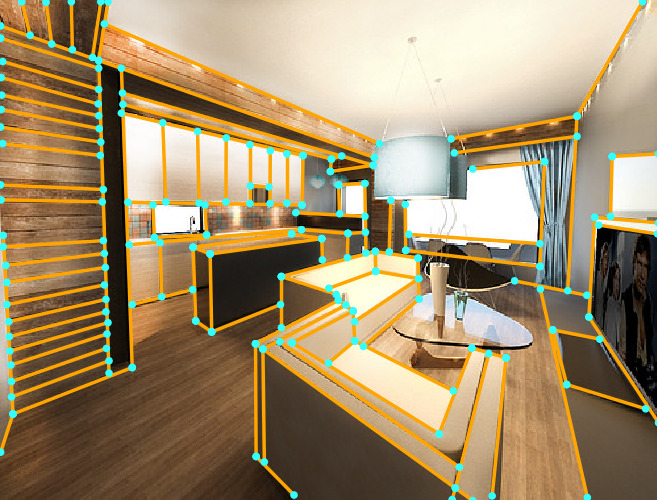}
    \includegraphics[width=0.99\linewidth]{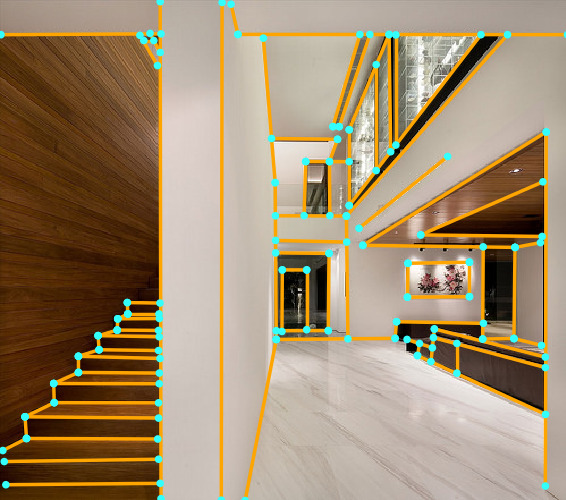}
    \includegraphics[width=0.99\linewidth]{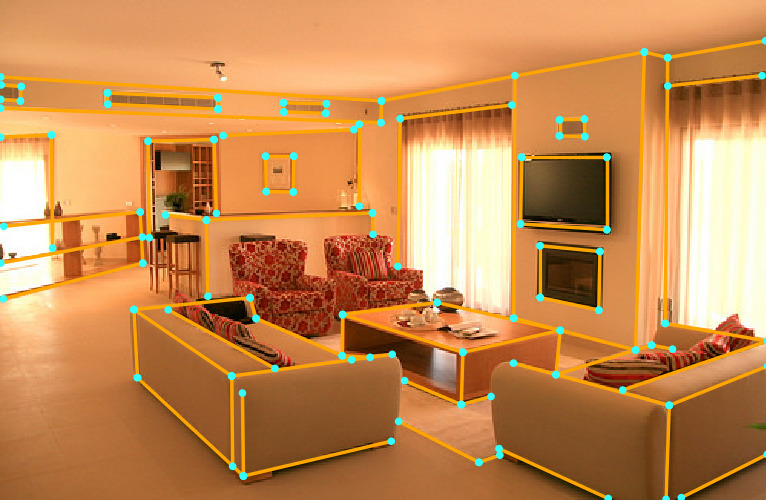}
    \includegraphics[width=0.99\linewidth]{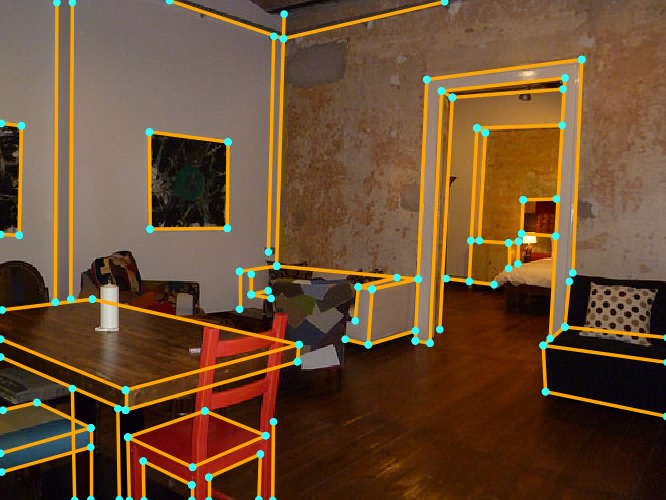}
    \includegraphics[width=0.99\linewidth]{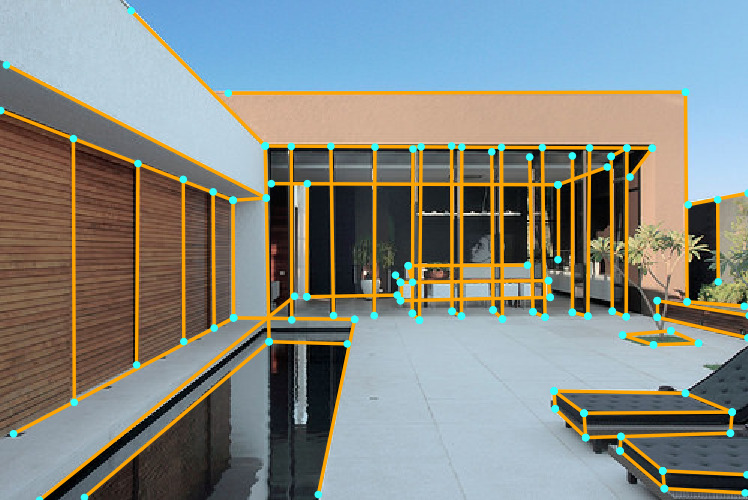}
    
    (e) Ground truth
    \end{minipage}\hfill
    
    \caption{Qualitative evaluation of wireframe and line detection methods.  From left to right, the columns shows the results from LSD \cite{von2010lsd}, AFM \cite{xue2018learning}, Wireframe \cite{Huang:2018:LPW}, L-CNN (ours), and the ground truth.  We also draw the detected junctions from Wireframe and L-CNN and the line endpoints from LSD and AFM.
    }
    \label{fig:qualitative}
\end{figure*}

We visualize our algorithm's output in \Cref{fig:qualitative}. The junctions are marked cyan blue and lines are marked orange.  Wireframes from L-CNN are post processed using the method from \N{Appendix A.1}.  Since LSD and AFM do not explicitly output junctions, we treat the endpoints of lines as junctions. As shown in \Cref{fig:qualitative}, LSD detects some high-frequency textures without semantic meaning.  This is expected as LSD is not a data-driven method.  By training a CNN to predict line heat maps, AFM \cite{xue2018learning} is able to suppress some noise.  However, both LSD and AFM still produce a lot of short lines because they do not have an explicit notion of junctions.  The wireframe parser \cite{Huang:2018:LPW} utilizes junctions to provide a relatively cleaner result, but their heuristic vectorization algorithm is sub-optimal and leads to crossing lines and incorrectly connected junctions. In contrast, our L-CNN uses powerful neural networks to classify whether a line indeed exists and thus provides the best performance.

\section*{Acknowledgement}
This work is partially supported by research grants from Sony US Research Center and Berkeley BAIR. We thank Kenji Tashiro of Sony for helpful discussions. We thank Cecilia Zhang of Berkeley for her helpful comments on the draft of this paper.  

\section{Supplementary Material}

\subsection{Post Processing} \label{sec:post}
Due to the existences of col-linear junctions in the scenes, we find that it is still common for L-CNN to generate overlapped lines.  Those overlapped lines are visually unpleasant and affect the quantitative performance metrics such as AP${}^H$. Therefore, we need to remove those overlapped lines via a post-processing stage.  Our strategy is inspired by the post processing techniques from object detection \cite{girshick2014rich}.  For each pair of lines $L_i=\{(\tilde \junction^1_i, \tilde \junction^2_i)\}$ and $L_j=\{(\tilde \junction^1_j, \tilde \junction^2_j)\}$ in which $L_i$ is ranked above $L_j$ according the line verification network, if $L_i$ is \emph{close} to $L_j$ (defined below), we
\begin{enumerate}[nosep]
    \item delete line $L_j$ if both the projections of points $\tilde \junction^1_j$ and $\tilde \junction^2_j$ to $L_i$ fall inside the segment $L_i$;
    \item cut line $L_j$ so that it does not overlap with $L_i$ if only one of the projections of points $\tilde \junction^1_j$ and $\tilde \junction^2_j$ to $L_i$ fall inside the segment $L_i$;
    \item retain line $L_j$ otherwise.
\end{enumerate}
We consider line segment $L_i$ close to $L_j$ if and only if
\begin{align*}
    \min(\max(&d(\tilde \junction^1_j, L_i), d(\tilde \junction^2_j, L_i)), \\
         \max(&d(\tilde \junction^1_i, L_j), d(\tilde \junction^2_i, L_j))) \le \eta_S,
\end{align*}
where $d(\junction, L)$ represents the distance between point $\junction$ and line segment $L$.  We set $\eta_S=0.01$ in our experiments.

Post processing improves visual appearance and performance when structural information is not that important. After post processing, $\text{F}^{\text{H}}$ of L-CNN is increased from 76.9 to 81.2 and $\text{AP}^{\text{H}}$ is increased from 80.3 to 82.8  as showned in \Cref{tab:result} on the ShanghaiTech dataset \cite{Huang:2018:LPW}. We do not post process the wireframes when the structural correctness is important (e.g. when evaluating sAP).

\subsection{Line Features}

\begin{table}[ht]
    \centering
    \begin{tabular}{c|cc|ccc}
             & cood & slope &  sAP${}^{5}$ & sAP${}^{10}$ & sAP${}^{15}$ \\ \hline\hline
         (a) &&& \textbf{58.9} & \textbf{62.9} & \textbf{64.7} \\
         (b) & \checkmark && 57.9 & 61.9 & 63.7 \\
         (c) & & \checkmark & 58.7 & 62.5 & 64.3 \\
         (d) & \checkmark & \checkmark & 58.0 & 61.9 & 63.7 \\
    \end{tabular}
    \caption{Ablation study of the coordinate-based manual line feature.  The column labelled with ``cood'' represents whether the first four dimension of $\mathbf{F}$, the coordinates of lines' endpoints, are used as features, and the column labelled with ``slope`` represents whether the last two dimensions of $\mathbf{F}$, the slope of the lines, are used as features.}
    \label{tab:ffeature}
\end{table}

Besides the features from the LoIPool layers, we also design and test the manual feature derived from the coordinates of lines' endpoints.  For each line $L_j=(\tilde \junction^1_j, \tilde \junction^2_j)$, let the 6-dimension line feature vector be
\begin{equation*}
    \mathbf{F}_j = \begin{bmatrix}
      \tilde \junction^{1T}_j & \tilde \junction^{2T}_j & \left(\frac{\tilde \junction^1_j-\tilde \junction^2_j}{\|\tilde \junction^1_j-\tilde \junction^2_j\|_2}\right)^T \requireJunctype{& \tilde t_j^1 & \tilde t_j^2}
    \end{bmatrix},
\end{equation*}
in which the first four dimensions store the coordinates of its endpoints and the last two dimensions store its normalized line directions.  According to \Cref{tab:ffeature}, we find that those manual features do not really improve the performance.  This is probably because the features from LoIPool layers are already powerful enough for the line verification network.  We also observe that the validation loss starting increasing prematurely during the training when we add the coordinate-based feature, a phenomena indicating overfitting, so we do not include this feature into our final L-CNN.

\subsection{More Qualitative Results}

We show \textbf{randomly sampled} results from the wireframe dataset at the end of the paper.  From LSD \cite{von2010lsd}, Wireframe parser \cite{Huang:2018:LPW}, and AFM \cite{xue2018learning}, we use the same hyper-parameters and cut-off values as in \cite{xue2018learning}.  For our L-CNN, we display the predicted lines with scores greater than $0.98$.  The colors of the lines indicate their confidences.
\bibliographystyle{ieee_fullname}
\bibliography{main}

\onecolumn
\centering

\begin{minipage}[t]{0.19\linewidth}\centering LSD\end{minipage}\begin{minipage}[t]{0.19\linewidth}\centering Wireframe\end{minipage}\begin{minipage}[t]{0.19\linewidth}\centering AFM\end{minipage}\begin{minipage}[t]{0.19\linewidth}\centering L-CNN\end{minipage}\begin{minipage}[t]{0.19\linewidth}\centering GT\end{minipage}

\includegraphics[width=0.19\linewidth]{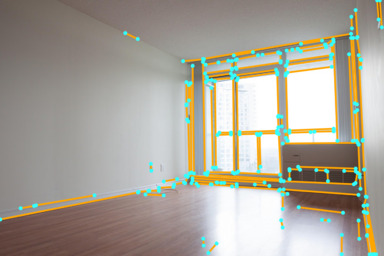}\includegraphics[width=0.19\linewidth]{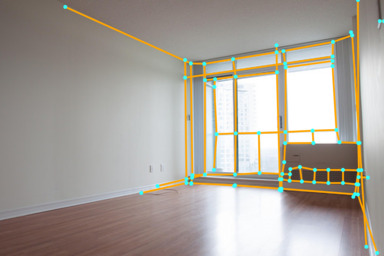}\includegraphics[width=0.19\linewidth]{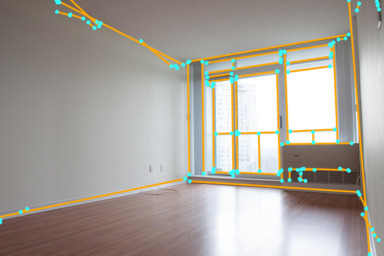}\includegraphics[width=0.19\linewidth]{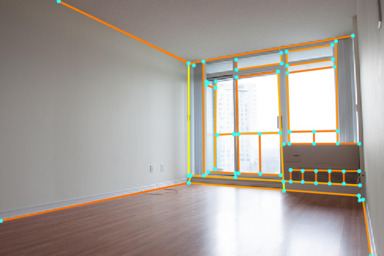}\includegraphics[width=0.19\linewidth]{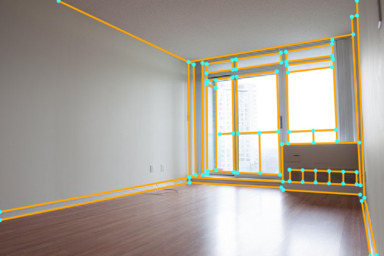}

\includegraphics[width=0.19\linewidth]{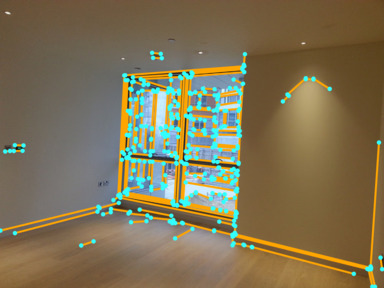}\includegraphics[width=0.19\linewidth]{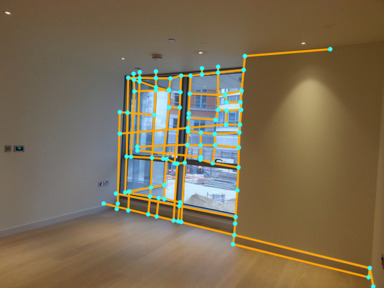}\includegraphics[width=0.19\linewidth]{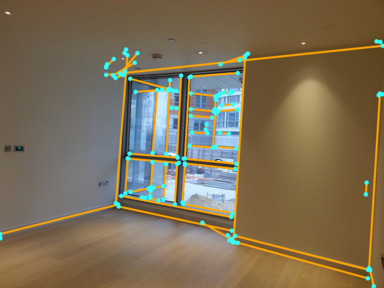}\includegraphics[width=0.19\linewidth]{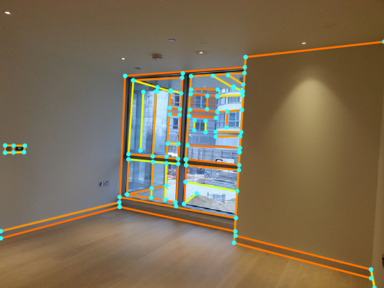}\includegraphics[width=0.19\linewidth]{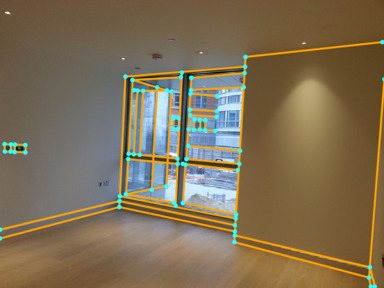}

\includegraphics[width=0.19\linewidth]{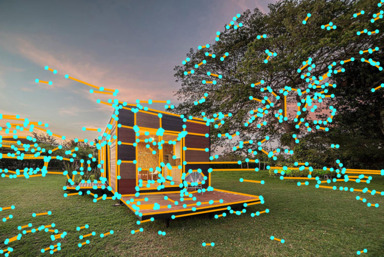}\includegraphics[width=0.19\linewidth]{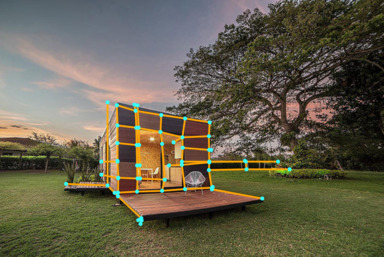}\includegraphics[width=0.19\linewidth]{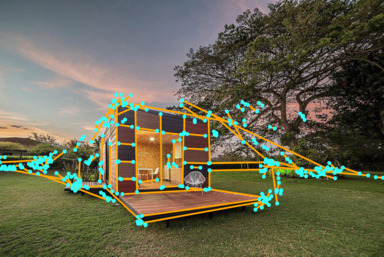}\includegraphics[width=0.19\linewidth]{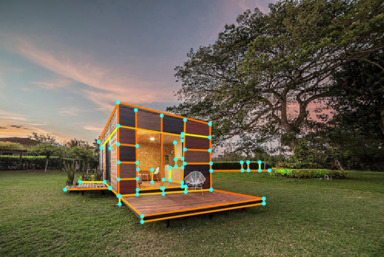}\includegraphics[width=0.19\linewidth]{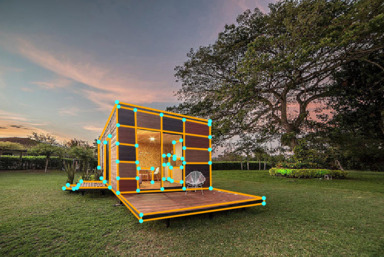}

\includegraphics[width=0.19\linewidth]{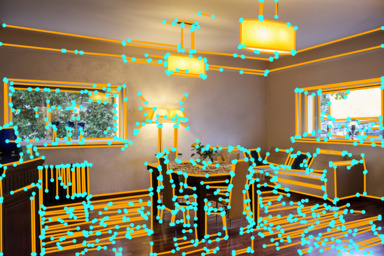}\includegraphics[width=0.19\linewidth]{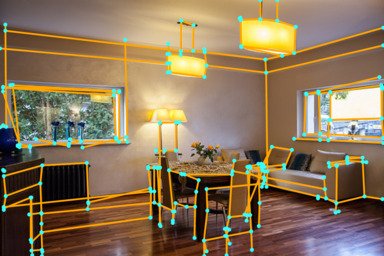}\includegraphics[width=0.19\linewidth]{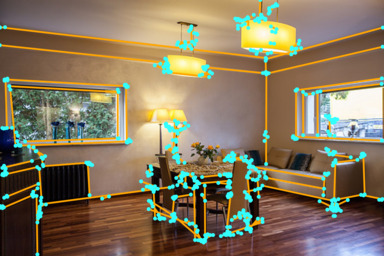}\includegraphics[width=0.19\linewidth]{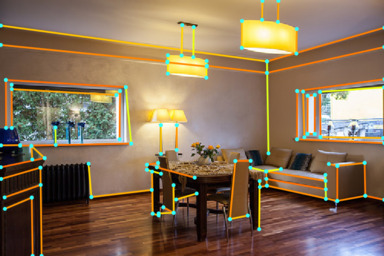}\includegraphics[width=0.19\linewidth]{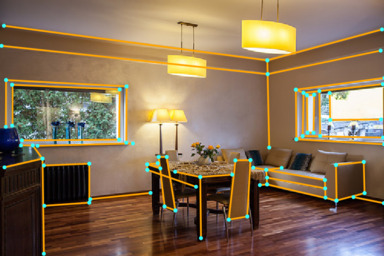}

\includegraphics[width=0.19\linewidth]{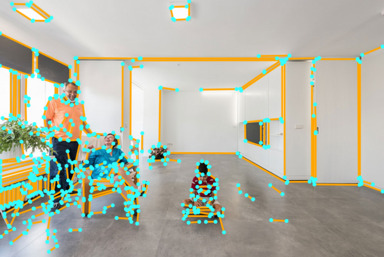}\includegraphics[width=0.19\linewidth]{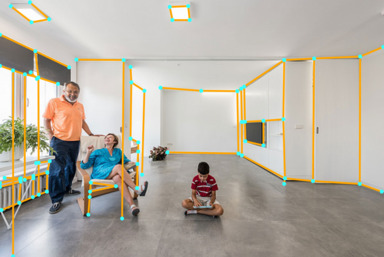}\includegraphics[width=0.19\linewidth]{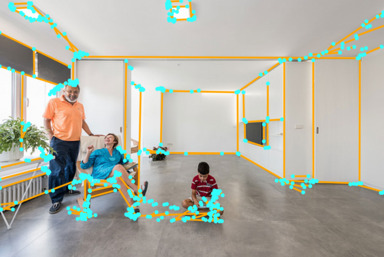}\includegraphics[width=0.19\linewidth]{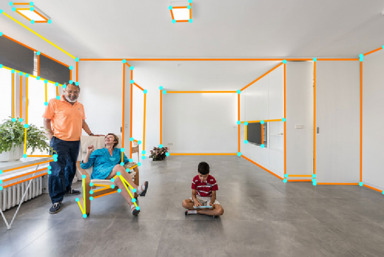}\includegraphics[width=0.19\linewidth]{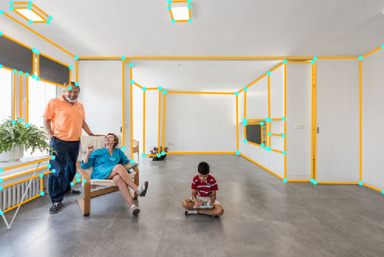}

\includegraphics[width=0.19\linewidth]{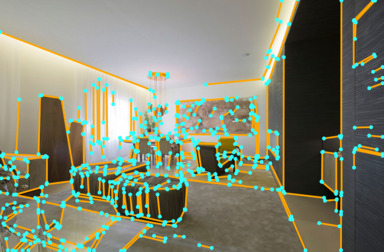}\includegraphics[width=0.19\linewidth]{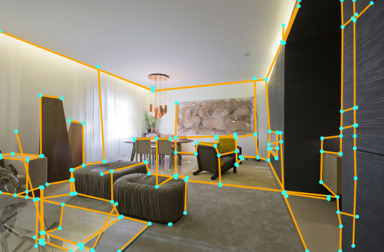}\includegraphics[width=0.19\linewidth]{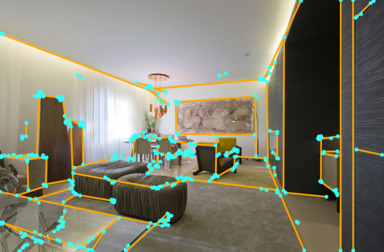}\includegraphics[width=0.19\linewidth]{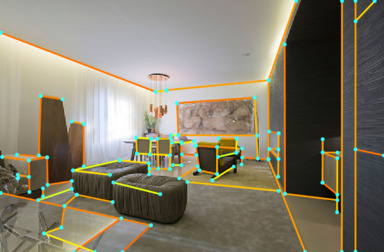}\includegraphics[width=0.19\linewidth]{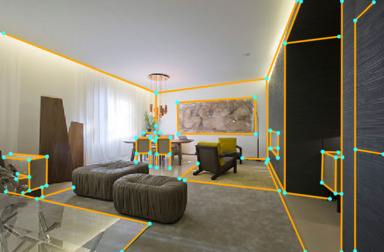}

\includegraphics[width=0.19\linewidth]{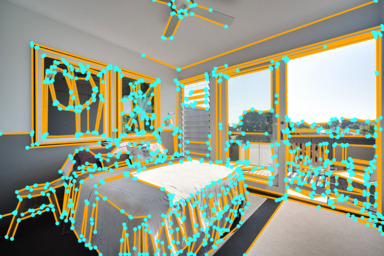}\includegraphics[width=0.19\linewidth]{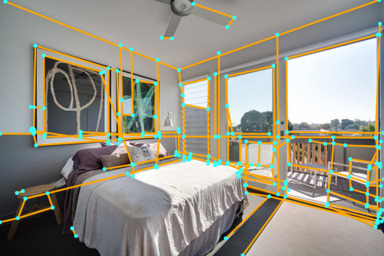}\includegraphics[width=0.19\linewidth]{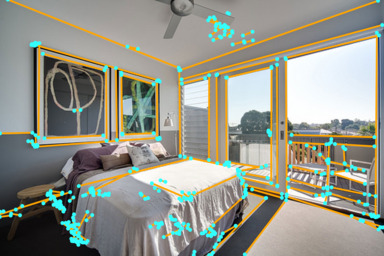}\includegraphics[width=0.19\linewidth]{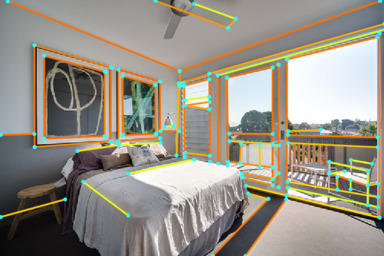}\includegraphics[width=0.19\linewidth]{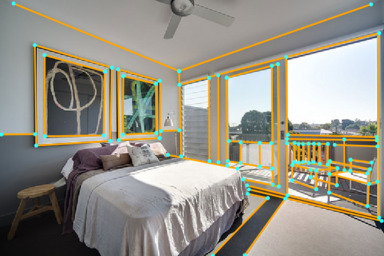}

\includegraphics[width=0.19\linewidth]{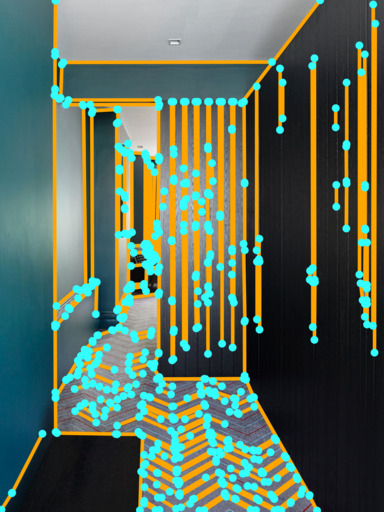}\includegraphics[width=0.19\linewidth]{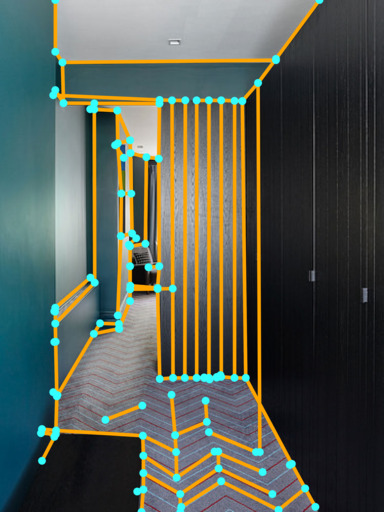}\includegraphics[width=0.19\linewidth]{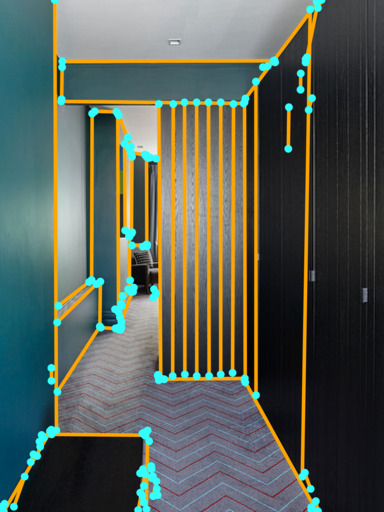}\includegraphics[width=0.19\linewidth]{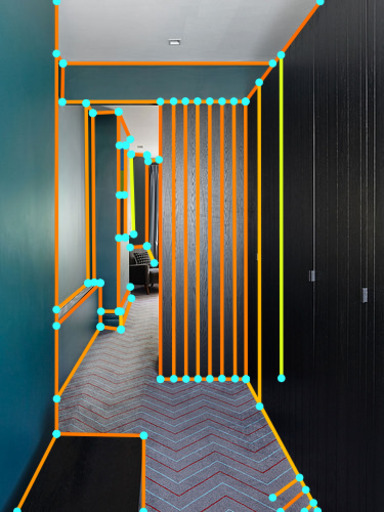}\includegraphics[width=0.19\linewidth]{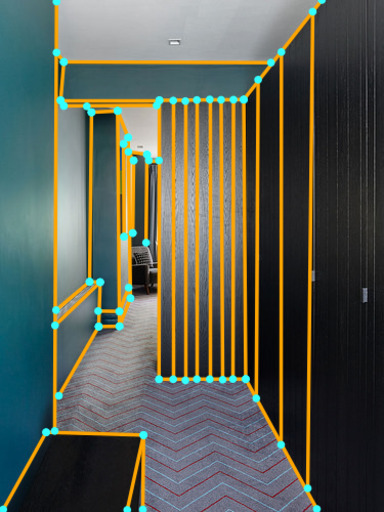}

\includegraphics[width=0.19\linewidth]{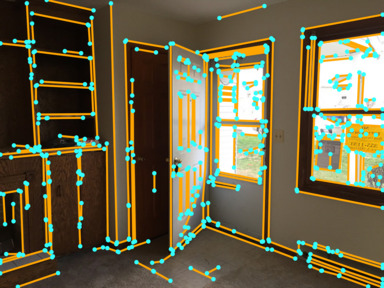}\includegraphics[width=0.19\linewidth]{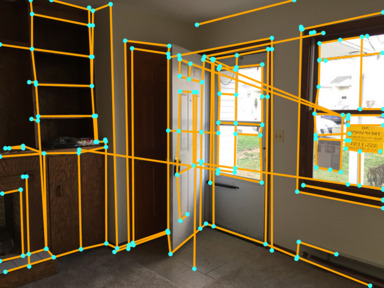}\includegraphics[width=0.19\linewidth]{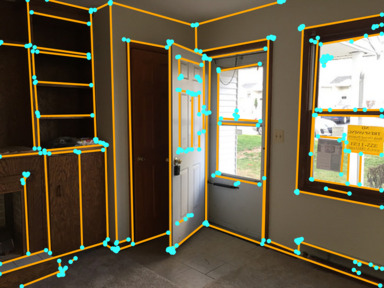}\includegraphics[width=0.19\linewidth]{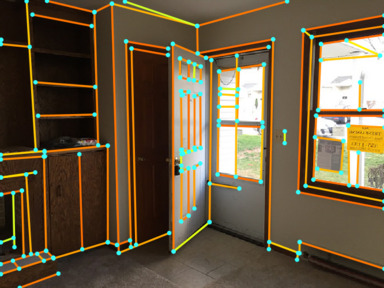}\includegraphics[width=0.19\linewidth]{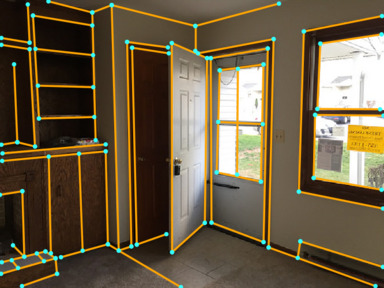}

\includegraphics[width=0.19\linewidth]{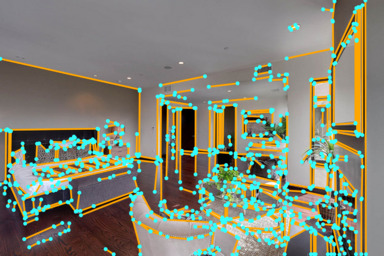}\includegraphics[width=0.19\linewidth]{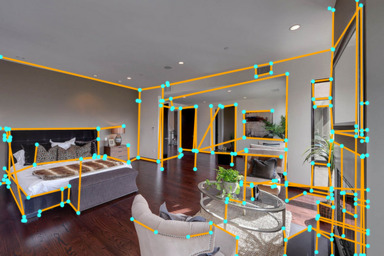}\includegraphics[width=0.19\linewidth]{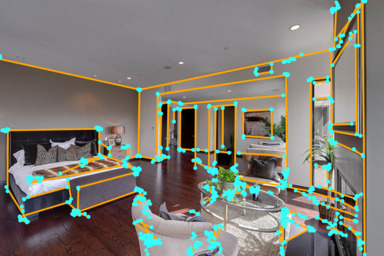}\includegraphics[width=0.19\linewidth]{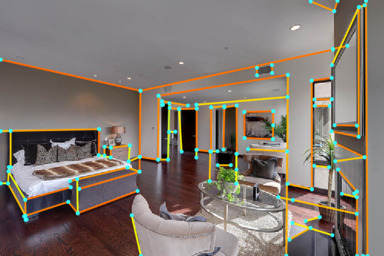}\includegraphics[width=0.19\linewidth]{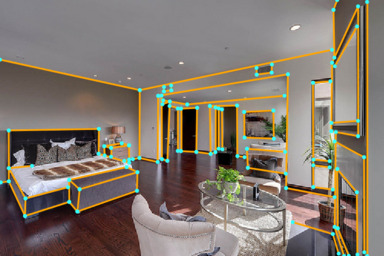}

\includegraphics[width=0.19\linewidth]{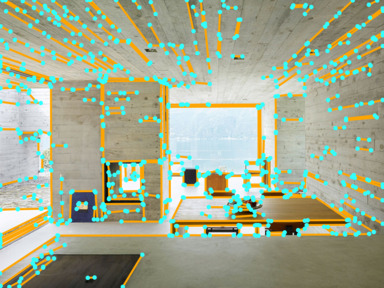}\includegraphics[width=0.19\linewidth]{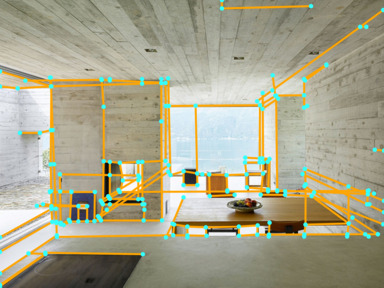}\includegraphics[width=0.19\linewidth]{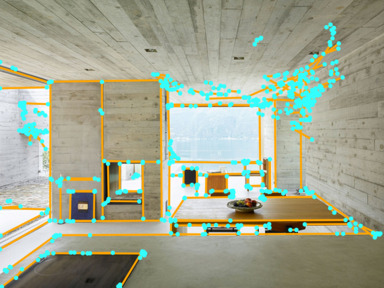}\includegraphics[width=0.19\linewidth]{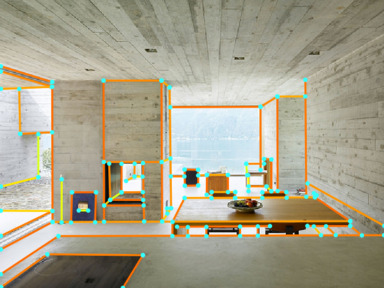}\includegraphics[width=0.19\linewidth]{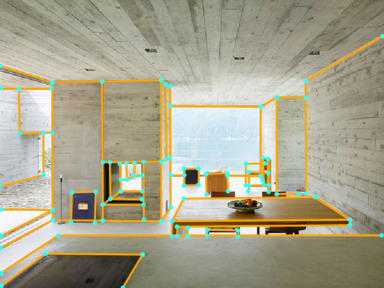}

\includegraphics[width=0.19\linewidth]{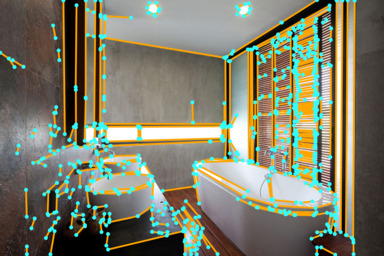}\includegraphics[width=0.19\linewidth]{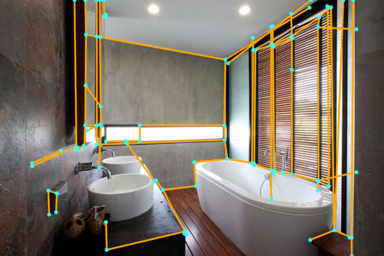}\includegraphics[width=0.19\linewidth]{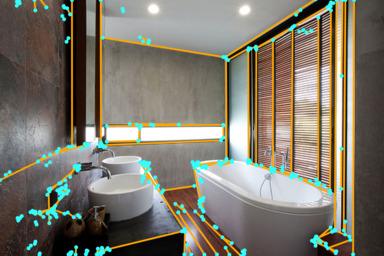}\includegraphics[width=0.19\linewidth]{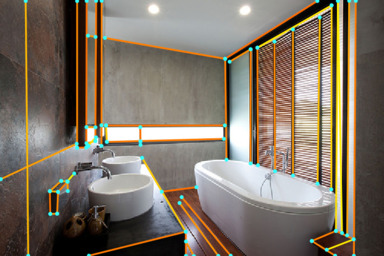}\includegraphics[width=0.19\linewidth]{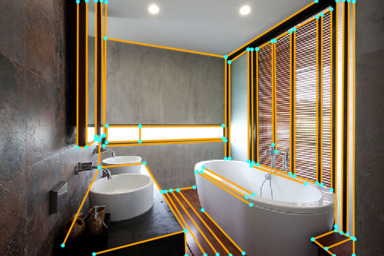}

\includegraphics[width=0.19\linewidth]{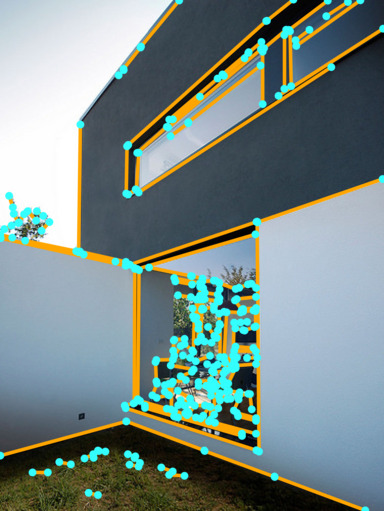}\includegraphics[width=0.19\linewidth]{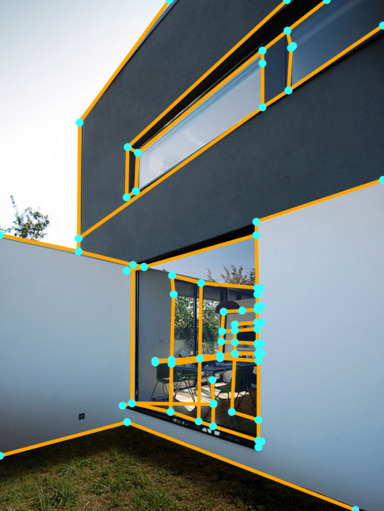}\includegraphics[width=0.19\linewidth]{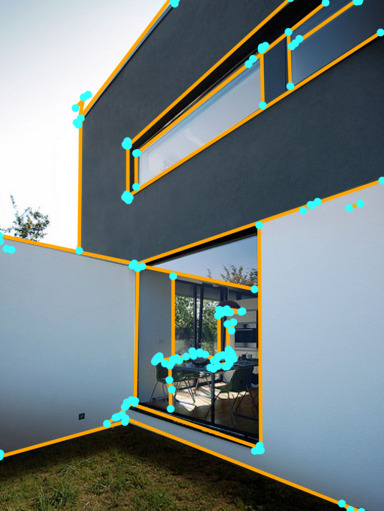}\includegraphics[width=0.19\linewidth]{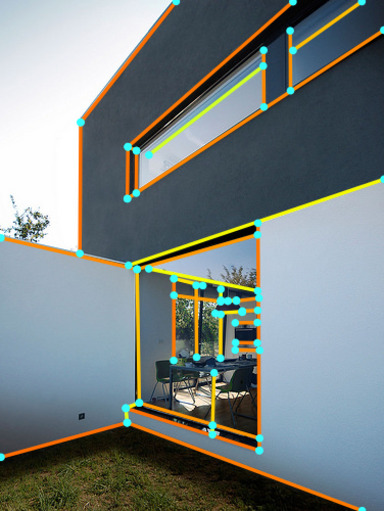}\includegraphics[width=0.19\linewidth]{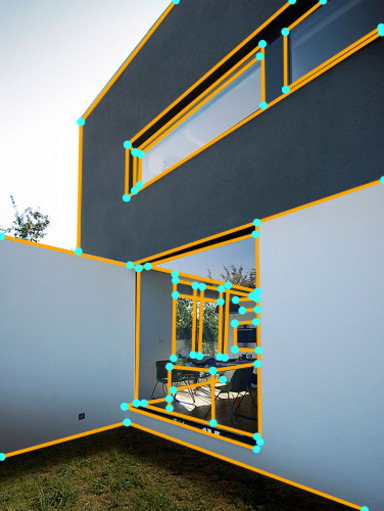}

\includegraphics[width=0.19\linewidth]{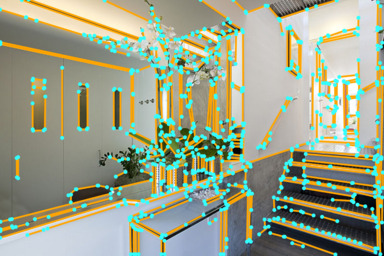}\includegraphics[width=0.19\linewidth]{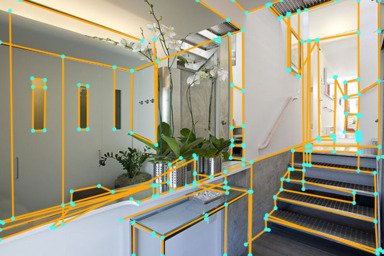}\includegraphics[width=0.19\linewidth]{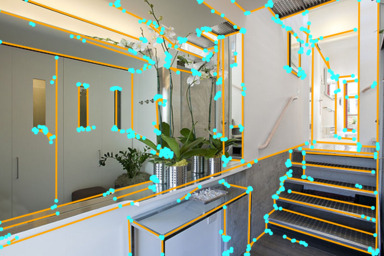}\includegraphics[width=0.19\linewidth]{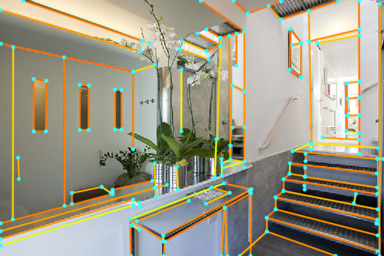}\includegraphics[width=0.19\linewidth]{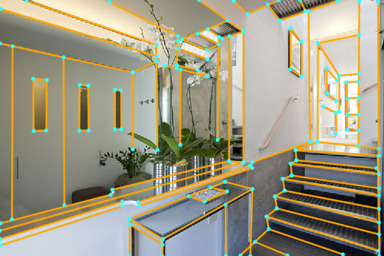}

\includegraphics[width=0.19\linewidth]{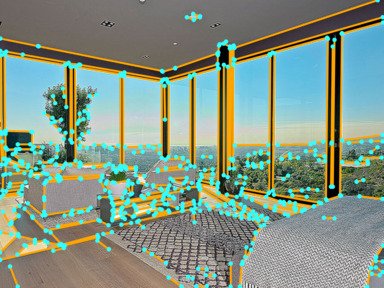}\includegraphics[width=0.19\linewidth]{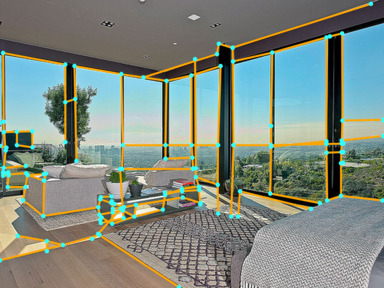}\includegraphics[width=0.19\linewidth]{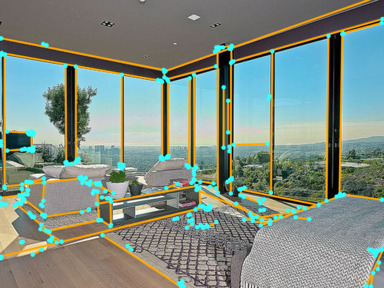}\includegraphics[width=0.19\linewidth]{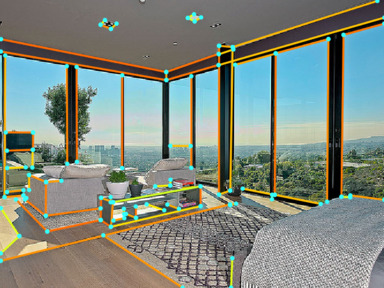}\includegraphics[width=0.19\linewidth]{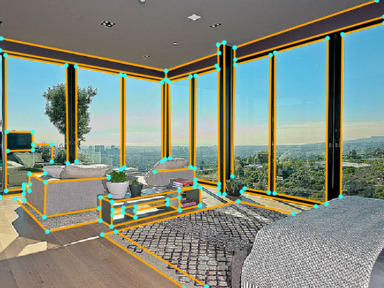}

\includegraphics[width=0.19\linewidth]{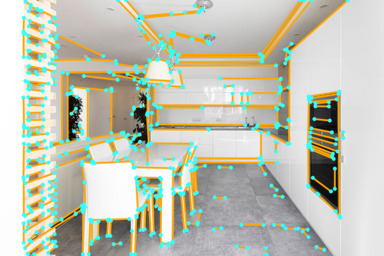}\includegraphics[width=0.19\linewidth]{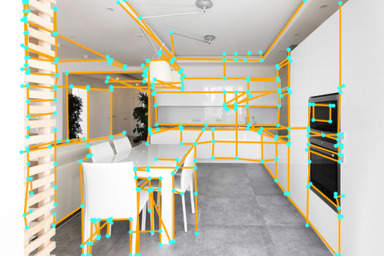}\includegraphics[width=0.19\linewidth]{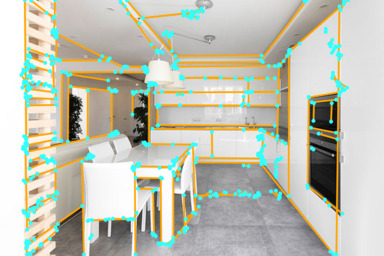}\includegraphics[width=0.19\linewidth]{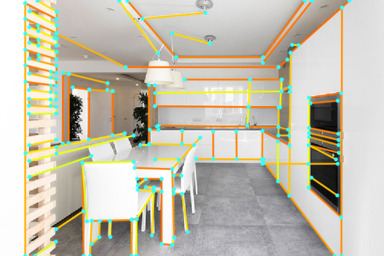}\includegraphics[width=0.19\linewidth]{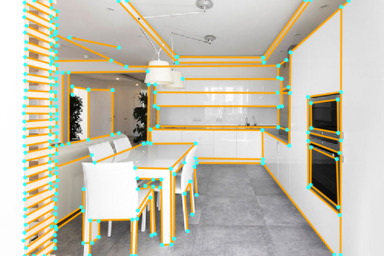}

\includegraphics[width=0.19\linewidth]{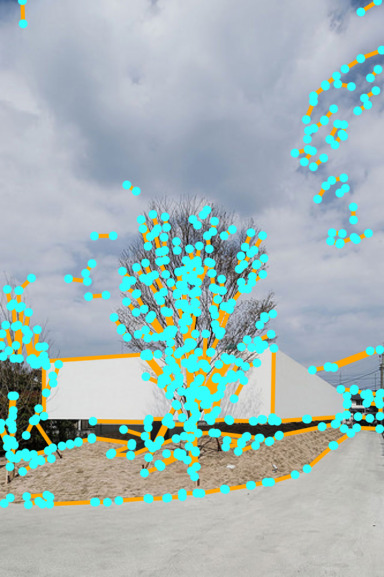}\includegraphics[width=0.19\linewidth]{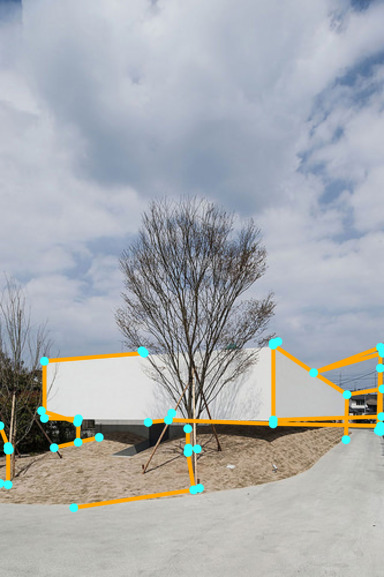}\includegraphics[width=0.19\linewidth]{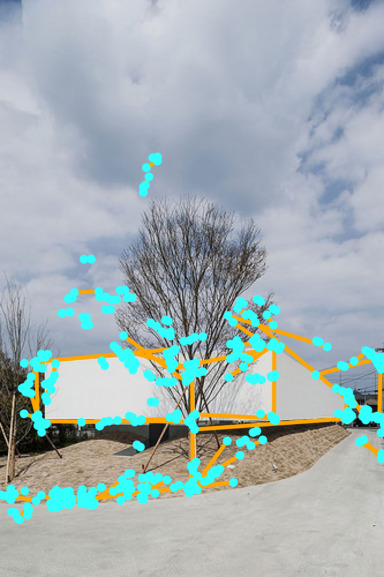}\includegraphics[width=0.19\linewidth]{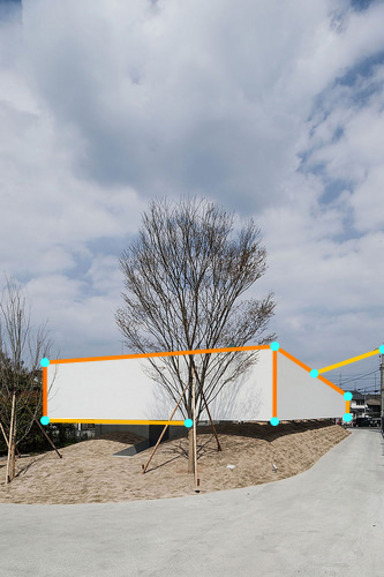}\includegraphics[width=0.19\linewidth]{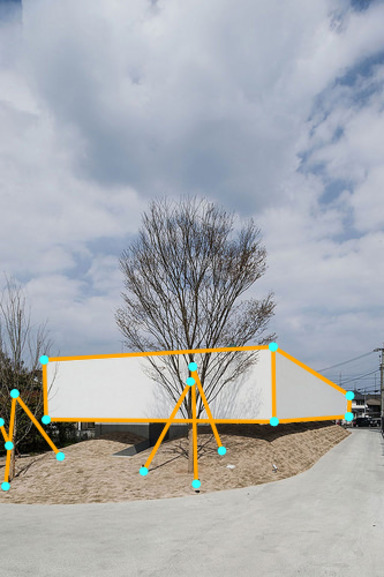}

\includegraphics[width=0.19\linewidth]{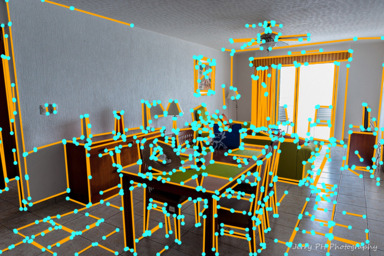}\includegraphics[width=0.19\linewidth]{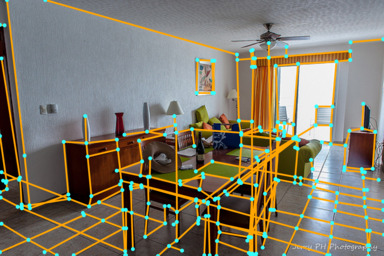}\includegraphics[width=0.19\linewidth]{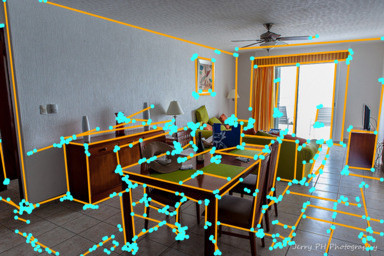}\includegraphics[width=0.19\linewidth]{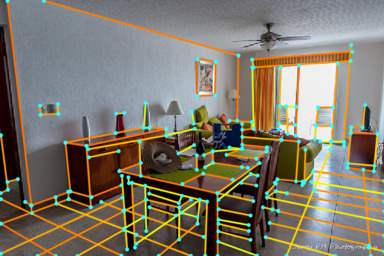}\includegraphics[width=0.19\linewidth]{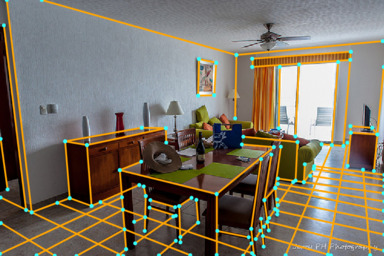}

\includegraphics[width=0.19\linewidth]{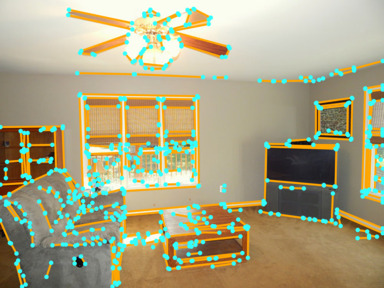}\includegraphics[width=0.19\linewidth]{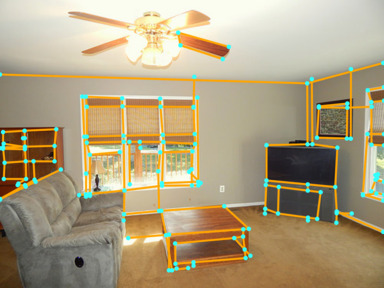}\includegraphics[width=0.19\linewidth]{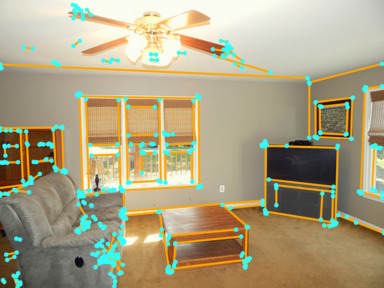}\includegraphics[width=0.19\linewidth]{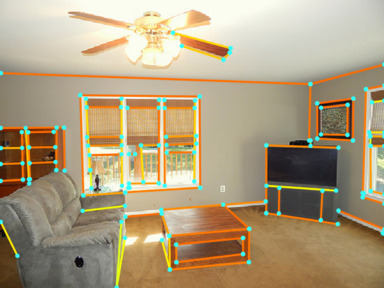}\includegraphics[width=0.19\linewidth]{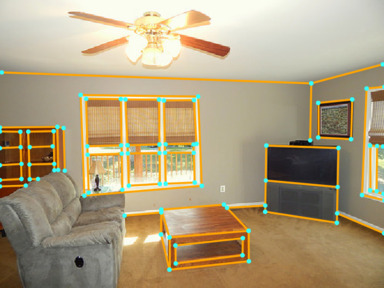}

\includegraphics[width=0.19\linewidth]{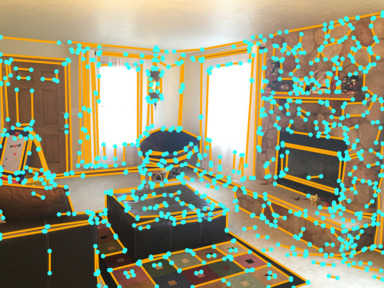}\includegraphics[width=0.19\linewidth]{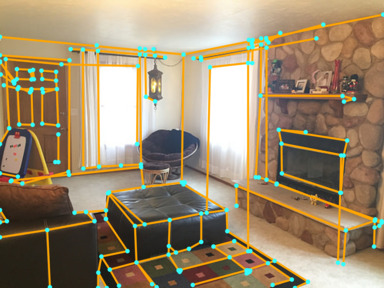}\includegraphics[width=0.19\linewidth]{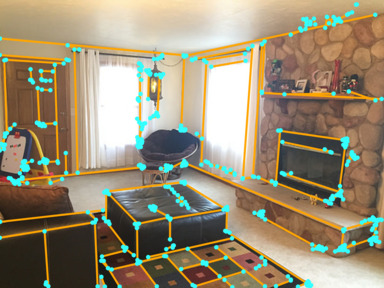}\includegraphics[width=0.19\linewidth]{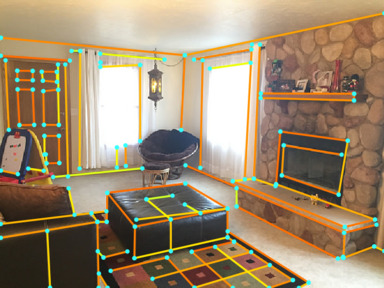}\includegraphics[width=0.19\linewidth]{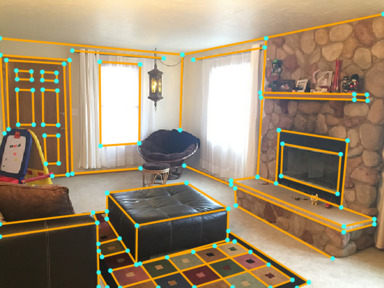}

\includegraphics[width=0.19\linewidth]{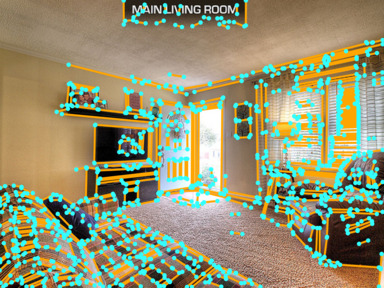}\includegraphics[width=0.19\linewidth]{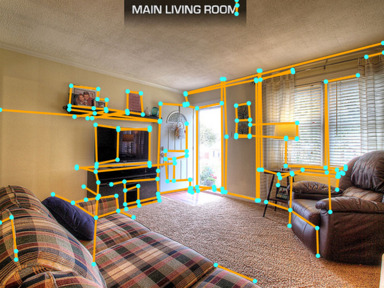}\includegraphics[width=0.19\linewidth]{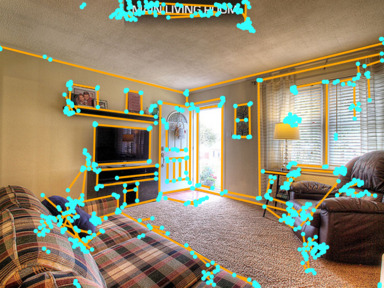}\includegraphics[width=0.19\linewidth]{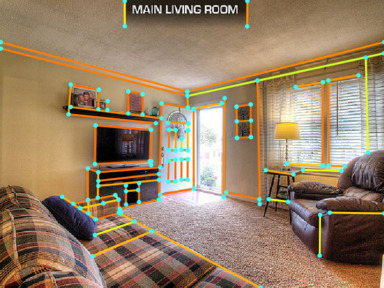}\includegraphics[width=0.19\linewidth]{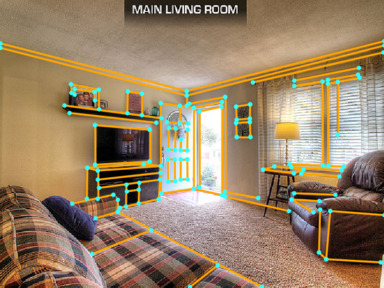}

\includegraphics[width=0.19\linewidth]{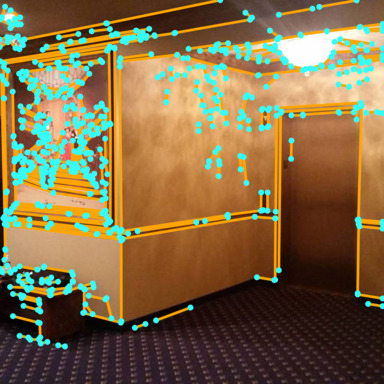}\includegraphics[width=0.19\linewidth]{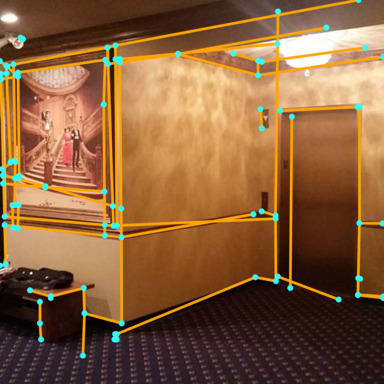}\includegraphics[width=0.19\linewidth]{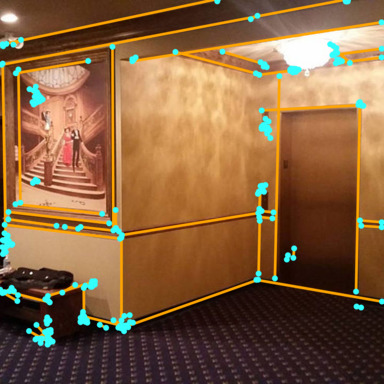}\includegraphics[width=0.19\linewidth]{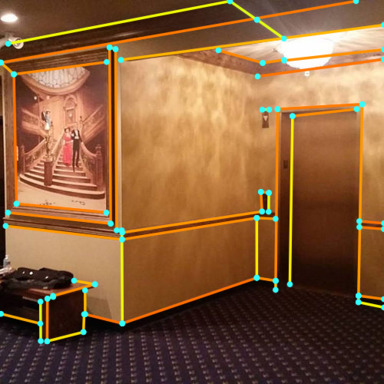}\includegraphics[width=0.19\linewidth]{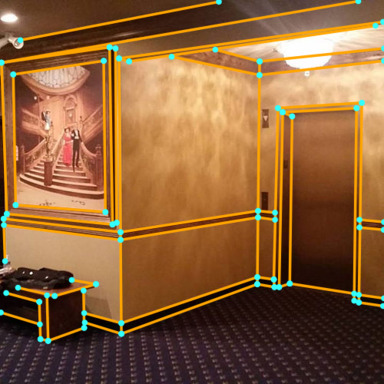}

\includegraphics[width=0.19\linewidth]{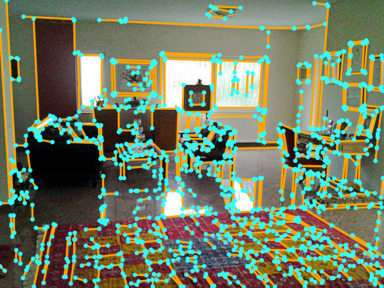}\includegraphics[width=0.19\linewidth]{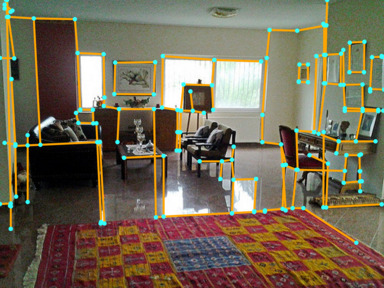}\includegraphics[width=0.19\linewidth]{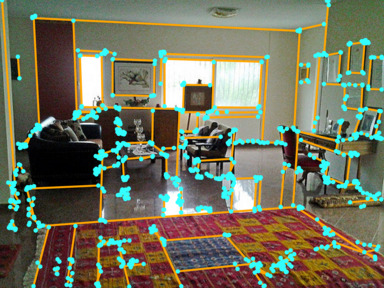}\includegraphics[width=0.19\linewidth]{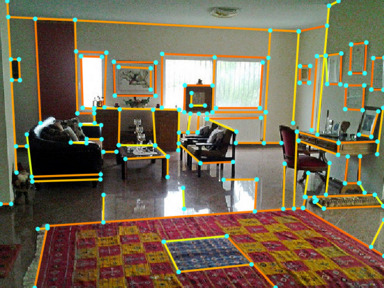}\includegraphics[width=0.19\linewidth]{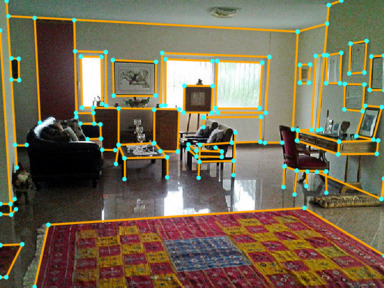}

\includegraphics[width=0.19\linewidth]{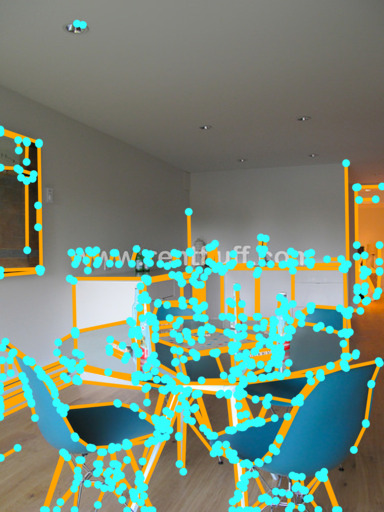}\includegraphics[width=0.19\linewidth]{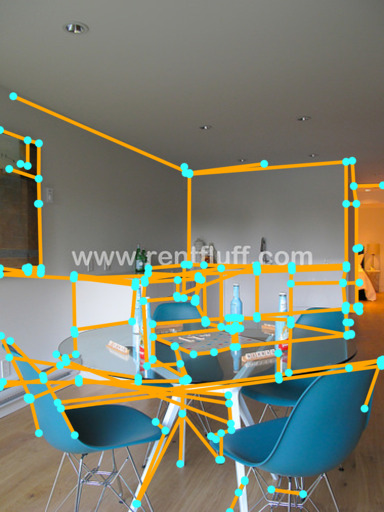}\includegraphics[width=0.19\linewidth]{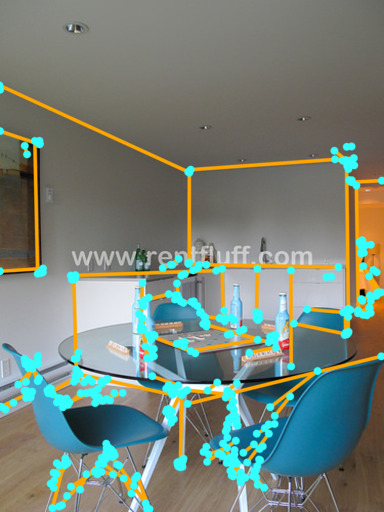}\includegraphics[width=0.19\linewidth]{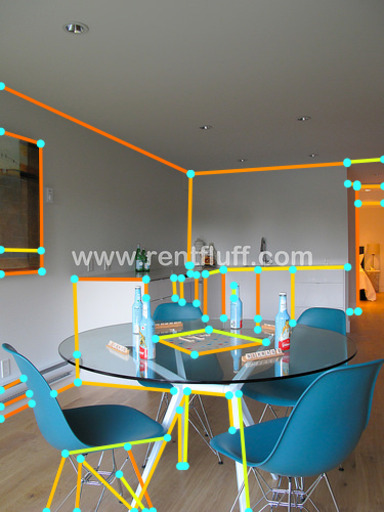}\includegraphics[width=0.19\linewidth]{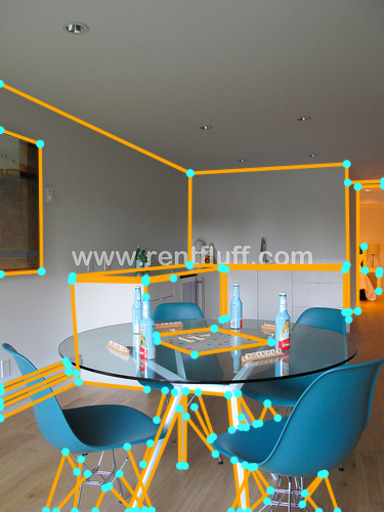}

\includegraphics[width=0.19\linewidth]{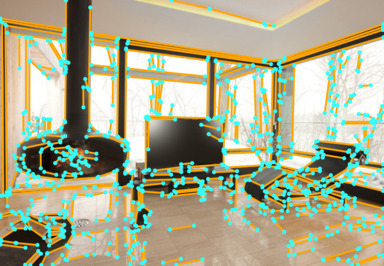}\includegraphics[width=0.19\linewidth]{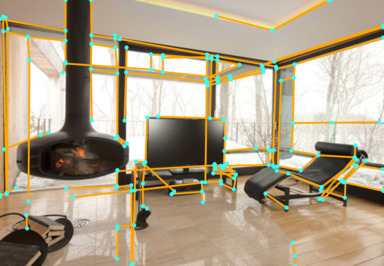}\includegraphics[width=0.19\linewidth]{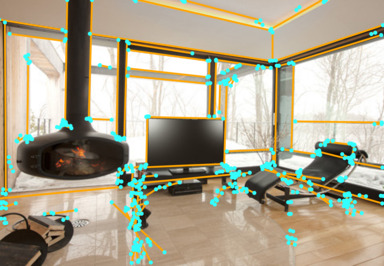}\includegraphics[width=0.19\linewidth]{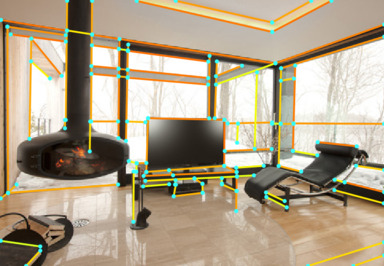}\includegraphics[width=0.19\linewidth]{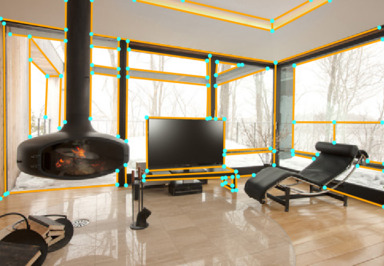}

\includegraphics[width=0.19\linewidth]{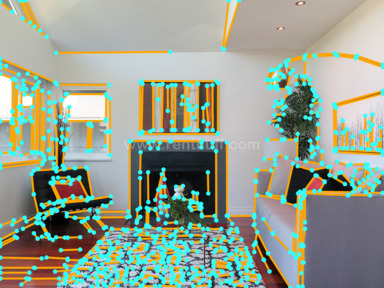}\includegraphics[width=0.19\linewidth]{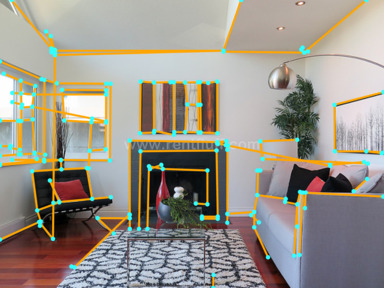}\includegraphics[width=0.19\linewidth]{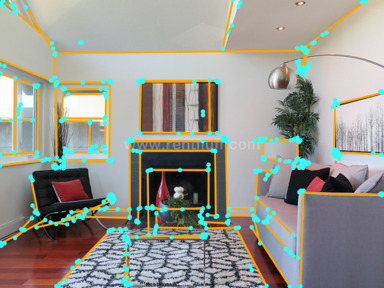}\includegraphics[width=0.19\linewidth]{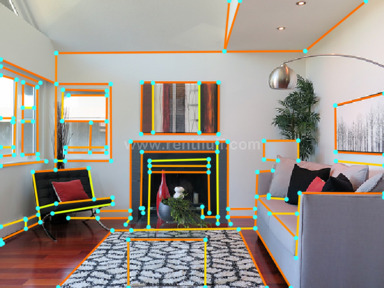}\includegraphics[width=0.19\linewidth]{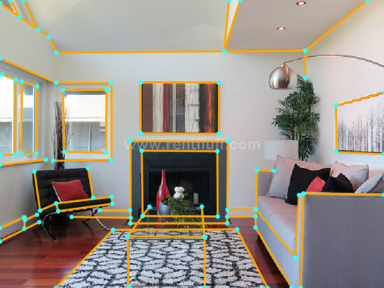}

\includegraphics[width=0.19\linewidth]{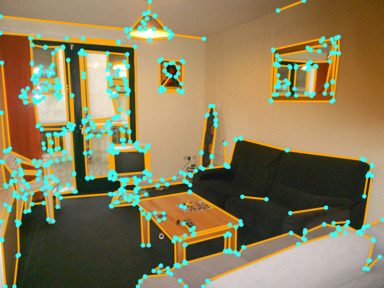}\includegraphics[width=0.19\linewidth]{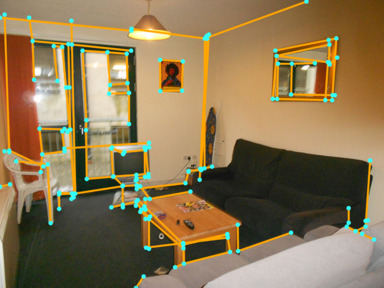}\includegraphics[width=0.19\linewidth]{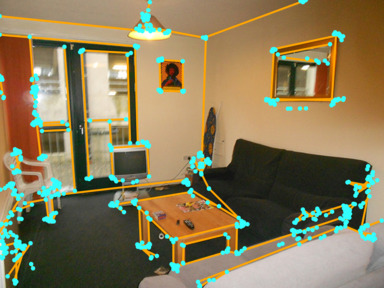}\includegraphics[width=0.19\linewidth]{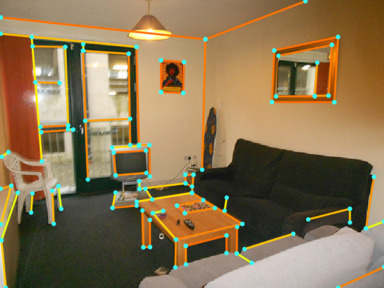}\includegraphics[width=0.19\linewidth]{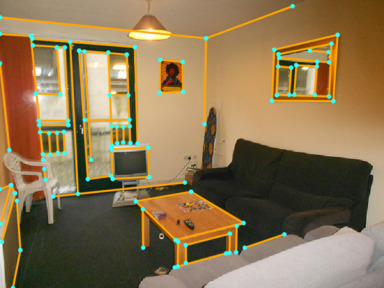}

\includegraphics[width=0.19\linewidth]{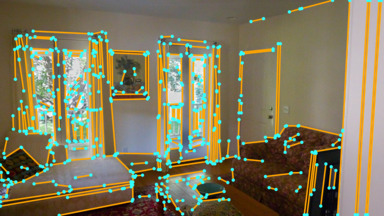}\includegraphics[width=0.19\linewidth]{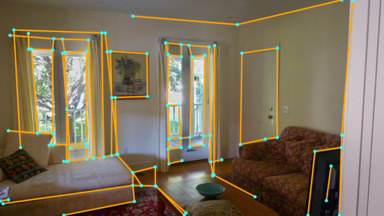}\includegraphics[width=0.19\linewidth]{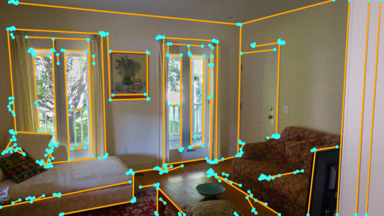}\includegraphics[width=0.19\linewidth]{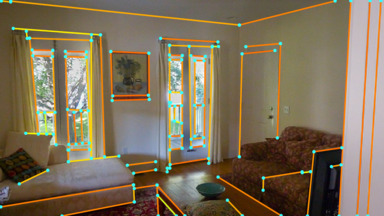}\includegraphics[width=0.19\linewidth]{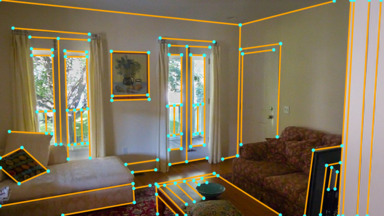}

\includegraphics[width=0.19\linewidth]{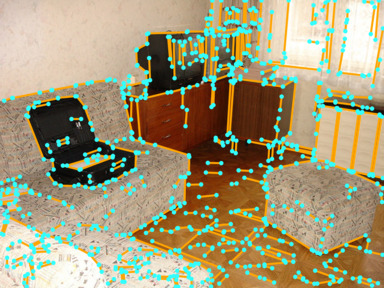}\includegraphics[width=0.19\linewidth]{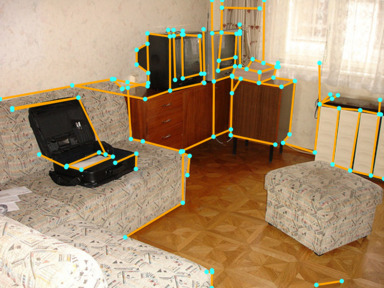}\includegraphics[width=0.19\linewidth]{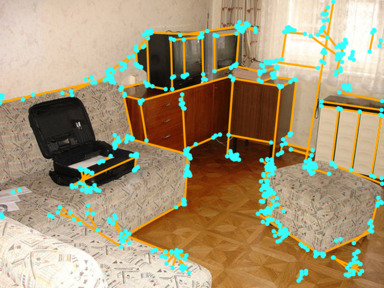}\includegraphics[width=0.19\linewidth]{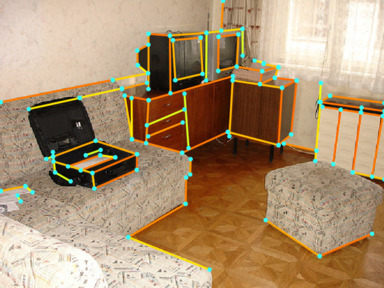}\includegraphics[width=0.19\linewidth]{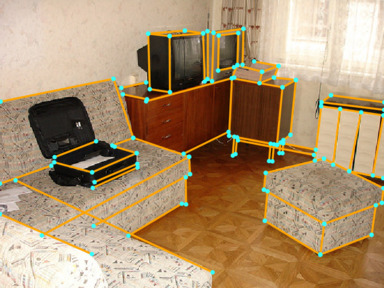}

\includegraphics[width=0.19\linewidth]{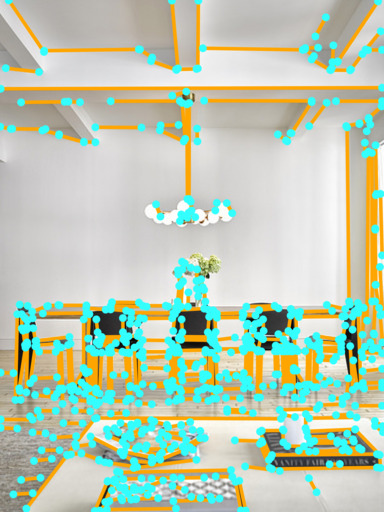}\includegraphics[width=0.19\linewidth]{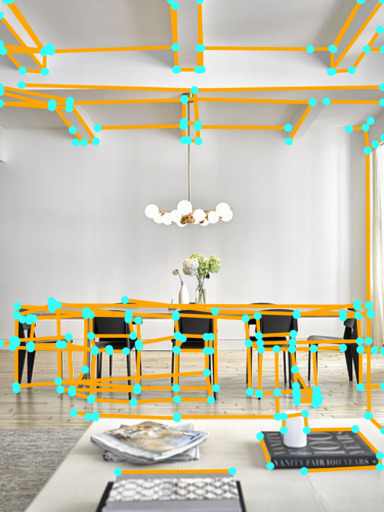}\includegraphics[width=0.19\linewidth]{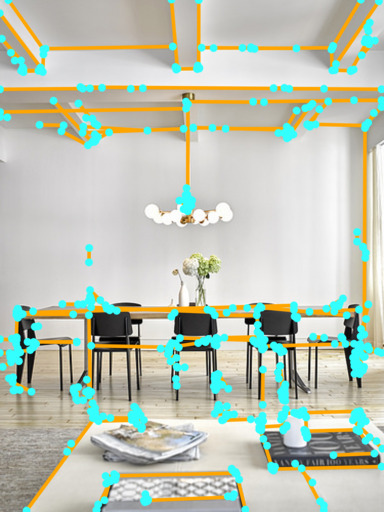}\includegraphics[width=0.19\linewidth]{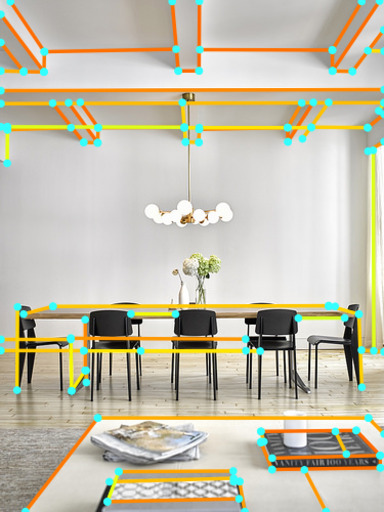}\includegraphics[width=0.19\linewidth]{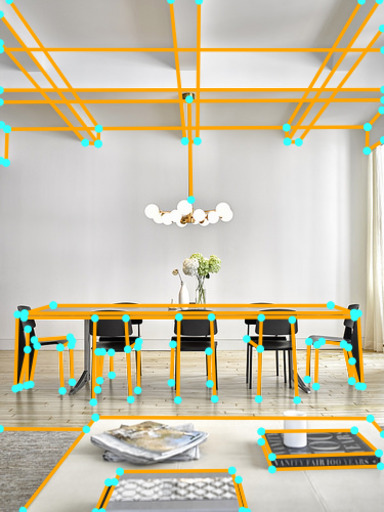}

\includegraphics[width=0.19\linewidth]{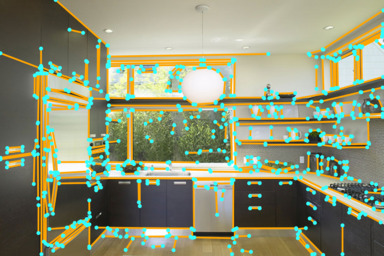}\includegraphics[width=0.19\linewidth]{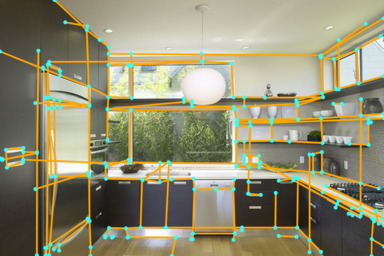}\includegraphics[width=0.19\linewidth]{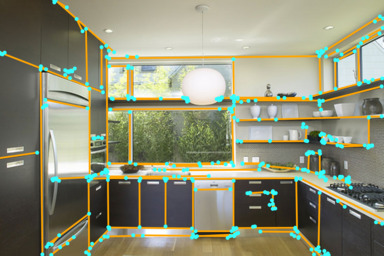}\includegraphics[width=0.19\linewidth]{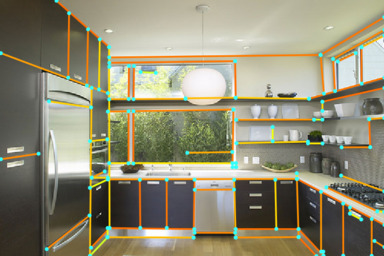}\includegraphics[width=0.19\linewidth]{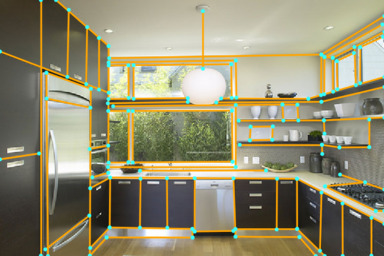}

\includegraphics[width=0.19\linewidth]{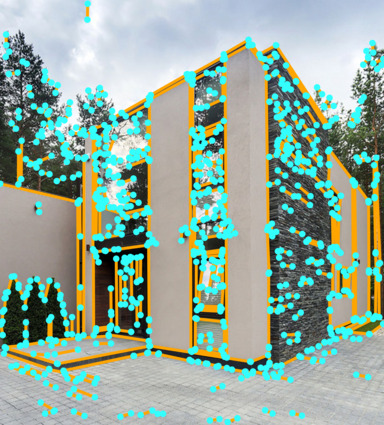}\includegraphics[width=0.19\linewidth]{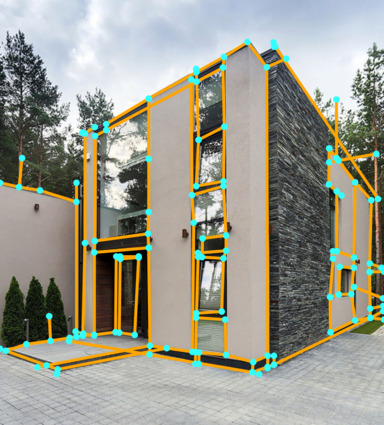}\includegraphics[width=0.19\linewidth]{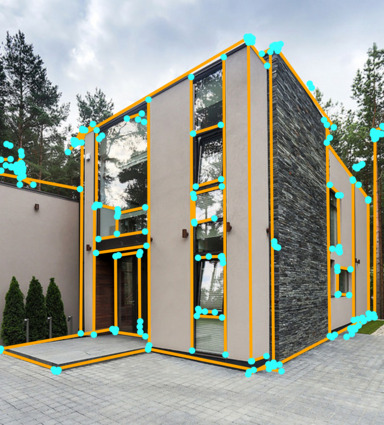}\includegraphics[width=0.19\linewidth]{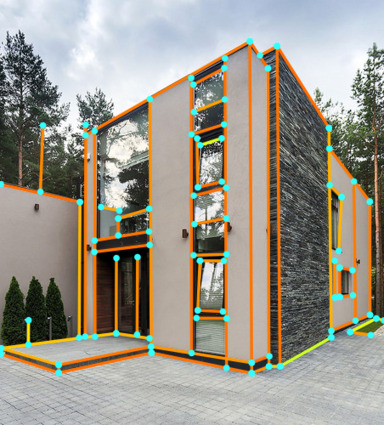}\includegraphics[width=0.19\linewidth]{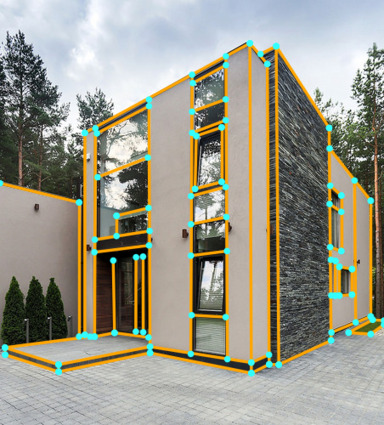}

\includegraphics[width=0.19\linewidth]{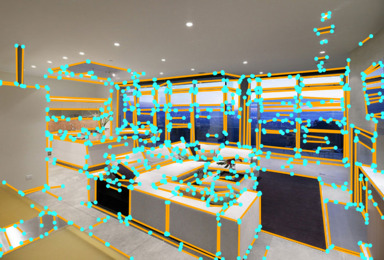}\includegraphics[width=0.19\linewidth]{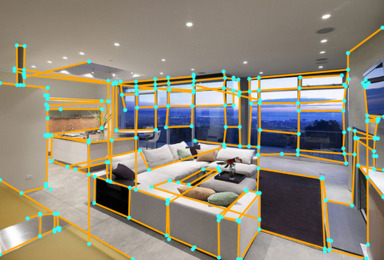}\includegraphics[width=0.19\linewidth]{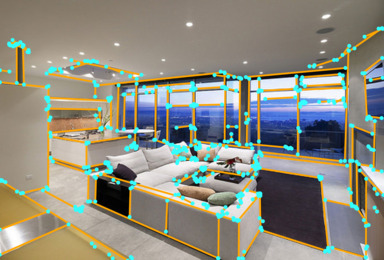}\includegraphics[width=0.19\linewidth]{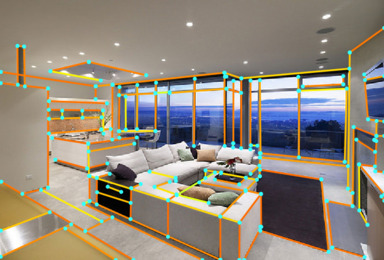}\includegraphics[width=0.19\linewidth]{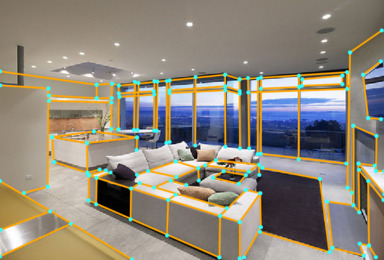}

\includegraphics[width=0.19\linewidth]{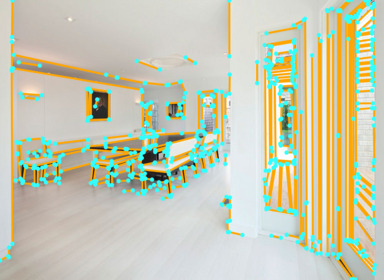}\includegraphics[width=0.19\linewidth]{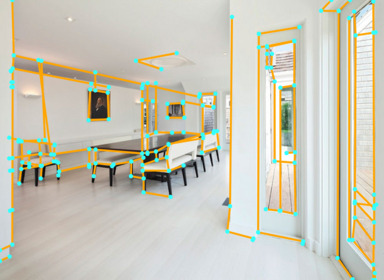}\includegraphics[width=0.19\linewidth]{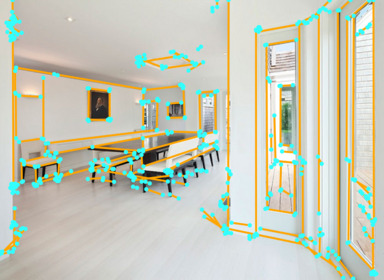}\includegraphics[width=0.19\linewidth]{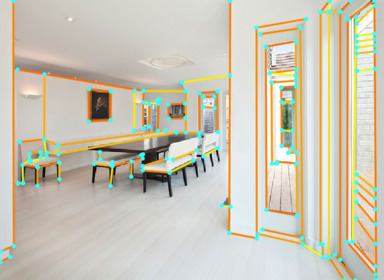}\includegraphics[width=0.19\linewidth]{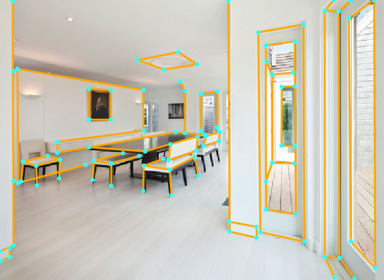}

\includegraphics[width=0.19\linewidth]{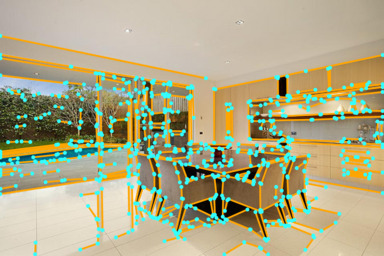}\includegraphics[width=0.19\linewidth]{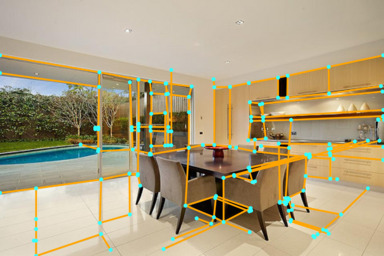}\includegraphics[width=0.19\linewidth]{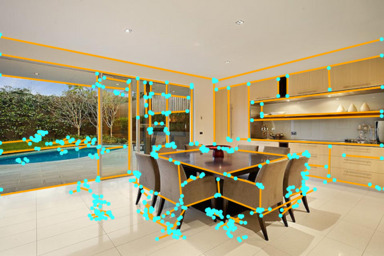}\includegraphics[width=0.19\linewidth]{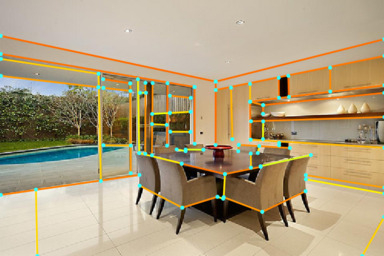}\includegraphics[width=0.19\linewidth]{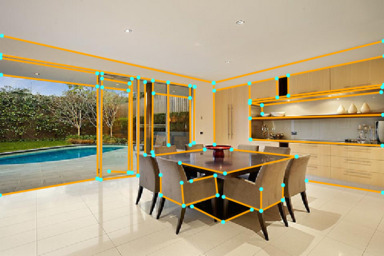}

\includegraphics[width=0.19\linewidth]{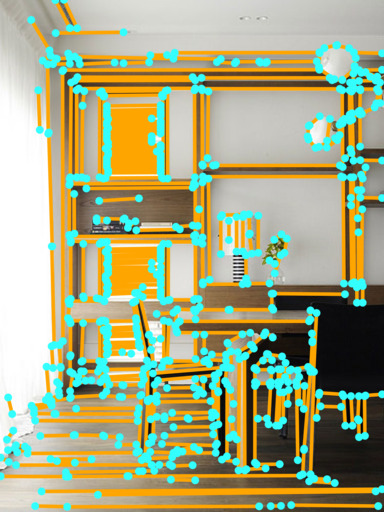}\includegraphics[width=0.19\linewidth]{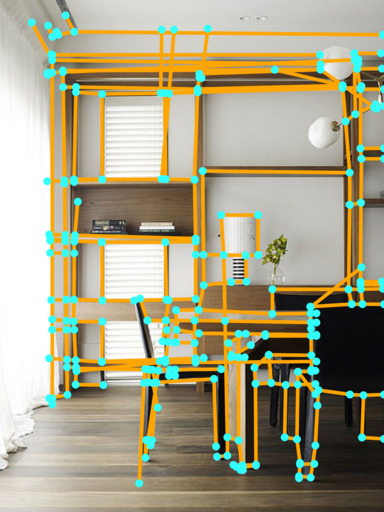}\includegraphics[width=0.19\linewidth]{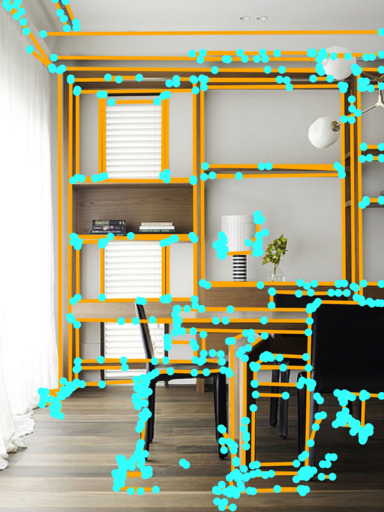}\includegraphics[width=0.19\linewidth]{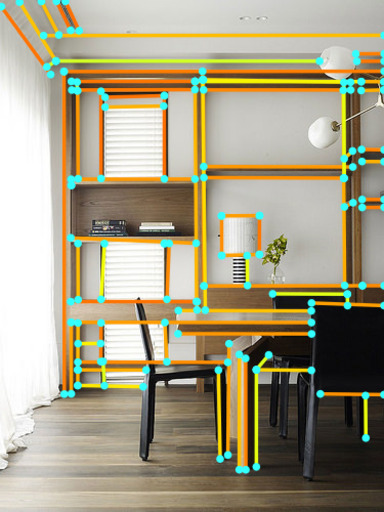}\includegraphics[width=0.19\linewidth]{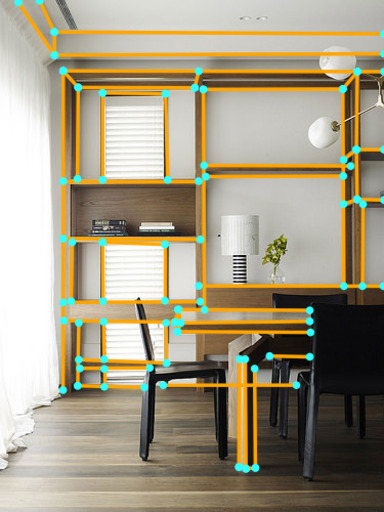}

\includegraphics[width=0.19\linewidth]{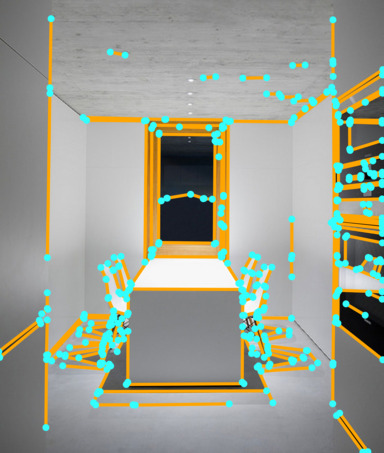}\includegraphics[width=0.19\linewidth]{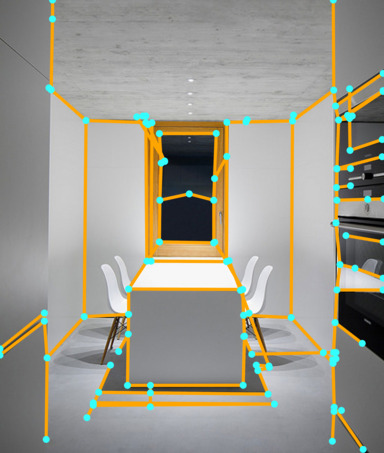}\includegraphics[width=0.19\linewidth]{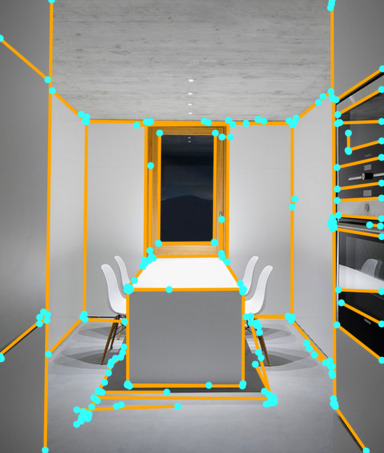}\includegraphics[width=0.19\linewidth]{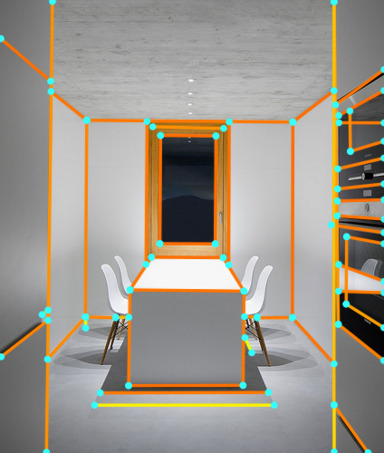}\includegraphics[width=0.19\linewidth]{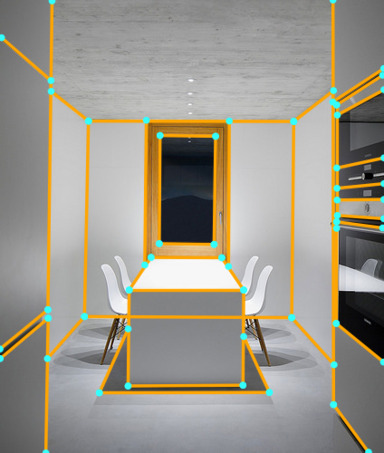}

\end{document}